%% file: main.tex
\documentclass[mathpazo]{cicp}
\newcommand{\tmmathbf}[1]{\ensuremath{\boldsymbol{#1}}}
\newcommand{\tmop}[1]{\ensuremath{\operatorname{#1}}}

\usepackage{algorithm,algorithmic}
\usepackage{xcolor}
\usepackage{subfigure}
\usepackage{graphicx}
\usepackage{booktabs}
\usepackage{makecell}
\begin{document}

\title{Reconstruction of dynamical systems without time label}



\author[F.~Author and A.~Co-Author(s)]{Zhijun Zeng\affil{1}, Chenglong Bao\affil{2}\comma\affil{3}, Pipi Hu\affil{4}, Yi Zhu\affil{2}\comma\affil{3}, Zuoqiang Shi\affil{2}\comma\affil{3}\comma\corrauth}

 \address{
 \affilnum{1}\ Department of Mathematical Sciences, Tsinghua University.\\
 \affilnum{2}\  Yau Mathematical Sciences Center, Tsinghua University.\\ 
 \affilnum{3}\ Yanqi Lake Beijing Institute of Mathematical Sciences and Applications.\\
  \affilnum{4}\ Microsoft Research AI4Science
  }

\emails{
{\tt zengzj22@mails.tsinghua.edu.cn} (F. Author),
{\tt zqshi@tsinghua.edu.cn} (A. Co-Author)
}

\begin{abstract}
 In this paper, we study the method to reconstruct dynamical systems from data without time labels. Data without time labels appear in many applications, such as molecular dynamics, single-cell RNA sequencing, etc. Reconstruction of dynamical system from time sequence data has been studied extensively. However, these methods do not apply if time labels are unknown. Without time labels, sequence data become distribution data. Based on this observation, we propose to treat the data as samples from a probability distribution and try to reconstruct the underlying dynamical system by minimizing the distribution loss, sliced Wasserstein distance more specifically. Extensive experiment results demonstrate the effectiveness of the proposed method. 
\end{abstract}

\maketitle
\input{introduction}
\input{problem}
\input{method}
\input{experiment}
\input{conclusion}
\bibliographystyle{abbrv}
\bibliography{references.bib}
\appendix
\input{appendix}
\end{document}

%% file: introduction.tex
\section{Introduction}\label{Intro}
Dynamical models are crucial for enhancing our comprehension of the natural world. By harnessing massive datasets to reveal the underlying governing equations that describe the behavior of complex physical systems, we can significantly advance our ability to model, simulate, and understand these systems across diverse scientific disciplines. It is a common situation that high-dimensional observations are generated by hidden dynamical systems operating within a low-dimensional space, a concept related to the manifold hypothesis.\cite{feffermanTestingManifoldHypothesis2016}. Accordingly, reconstructing the underlying dynamical system is crucial for simulating real-world scientific systems and educing its mechanisms. 

For an evolutionary system, the observations consist of trajectory data with time labels $\{(t_i,\boldsymbol{x}_i)\}_{i=1}^n$, typically modeled using a dynamic form $\boldsymbol{x}_{\boldsymbol{\theta}}(t)$ or its differential equation representation $\frac{d\boldsymbol{x}}{dt} = \boldsymbol{f}(\boldsymbol{x},\boldsymbol{\theta})$.  However, in certain contexts—such as microscopy—technical limitations preclude the acquisition of precise time labels, and the observed data are instead obtained as unlabeled high-dimensional point clouds $\{\boldsymbol{x}_i\}_{i=1}^n$. Although reconstructing the hidden dynamics from such unlabeled data remains challenging, one can still postulate reasonable assumptions regarding the observation times and attempt to reconstruct the dynamics.
\subsection*{System Identification} 
The conventional system identification task involves determining the form of the underlying system and estimating its parameters from trajectory data with time labels $\{(t_i,\boldsymbol{x}_i)\}_{i=1}^n$. Traditionally, the forward solver-based nonlinear least squares (FSNLS) method is a standard approach for system identification in models governed by differential equations, and its procedure can be summarized in four steps: (1) proposing an initial set of parameters, (2) solving the forward process on the collocation points using a numerical solver, (3) comparing the generated solution with observational samples and updating the parameters, and (4) iterating steps (2) and (3) until the convergence criteria are satisfied. As a well-established topic, \cite{banksEstimationTechniquesDistributed2012} and \cite{ljungSystemIdentification1998} provide a comprehensive overview of extant results. As models become larger and demand greater realism in FSNLS, two main challenges persist—solver accuracy and the selection of the initial candidate parameter—which significantly influence the final estimate, yet fully satisfactory solutions remain elusive. 
Recently, the Neural Ordinary Differential Equation (ODE) approach and its derivatives have combined FSNLS with a neural network ansatz for the forcing term and provided a powerful tool for inferring the complex dynamics of physical systems.\cite{zhongSymplecticODENetLearning2020}\cite{choiLearningQuantumDynamics2022}\cite{nakajimaNeuralSchrodingerEquation2022}\cite{huRevealingHiddenDynamics2020}. However, both FSNLS and Neural ODE update the model by computing pointwise residuals between simulated data and trajectory observations, which cannot be applied to unlabeled data.

To address these issue, an alternative approach , known as the sparse identification of nonlinear dynamics (SINDy)\cite{bruntonDiscoveringGoverningEquations2016}\cite{bellmanNewMethodIdentification1969}, avoids forward computation by evaluating the nonlinear candidate basis functions on the given dataset and estimates the system parameters using sparse regression methods. Numerous subsequent studies have augmented the performance of this foundational concept by refining the estimation of differentiation and least square\cite{messenger2021weak}\cite{kaheman2020sindy}\cite{fasel2022ensemble}. Both the strong-form and weak-form variants of SINDy rely on time label information to compute constraint equations, thereby precluding their direct application to unlabeled data. 
\subsection*{Extract dynamics from unlabeled data}
Unlabeled data refer to point cloud data that lack parameterization (e.g., observation time). Assuming these observational datasets are driven by an underlying dynamical system, we require additional information—the distribution of observation times—to reconstruct both the system and its missing parametrization, and the mathematical model of this problem is provided in Section~\ref{problem}.
For example, single-cell RNA sequencing (scRNA-seq) exemplifies the unlabeled data paradigm in genomics by collecting high-dimensional point cloud datasets. In these datasets, each observation represents an individual cell's gene expression profile and is drawn as an i.i.d. random sample from the distribution of high-dimensional cell expression over the finite interval $[0,T]$. To account for this, existing numerical methods model the ODEs governing cellular evolution and estimate model parameters by minimizing distributional distances (e.g., the Kullback–Leibler divergence). However, for general unlabeled data—where the underlying dynamics remain unknown and the observation time distribution is not necessarily uniform—this estimation approach is inapplicable. 

On one hand, disregarding the alignment with the observation time distribution allows pseudotemporal ordering to be viewed as extracting a one-dimensional parametrization of a manifold. Consequently, early trajectory inference studies leveraged manifold learning techniques—employing methods that preserve either global \cite{tenenbaumGlobalGeometricFramework2000,mcinnesUmapUniformManifold2018} or local data\cite{roweisNonlinearDimensionalityReduction2000,belkinLaplacianEigenmapsDimensionality2003,vandermaatenVisualizingDataUsing2008} structures to derive pseudotemporal orderings. More recent approaches have adopted graph-based methods \cite{grun2015single, wolfPAGAGraphAbstraction2019} and deep learning techniques \cite{du2020model} to eliminate the need for prior knowledge of network topology and enhance robustness.

On the other hand, one can employ generative models to construct the transport map between observation time and high-dimensional data points. In this field, numerous models have been developed to extract latent representations and derive such transport maps for generating real-world high-dimensional data—examples include variational autoencoders \cite{kingmaAutoencodingVariationalBayes2013}, GANs \cite{goodfellowGenerativeAdversarialNets2014}, normalizing flows \cite{kobyzevNormalizingFlowsIntroduction2020}, and diffusion models \cite{songScorebasedGenerativeModeling2020, vahdatScorebasedGenerativeModeling2021}. However, it is difficult to recover the underlying scientific principles from the resulting black-box models.

\subsection*{Contribution}
The main contribution of our paper is that we propose a framework that automates the reconstruction of time labels and infers the underlying hidden dynamics of observed trajectories. First, we build the mathematical model of inferring dynamics from unlabeled data using the Wasserstein metric. We then formulate a forward solver–based approach to solve this problem; however, this conventional framework is hampered by nonconvexity and numerical challenges associated with distributional metrics. To address these issues, 
we propose directly approximating the hidden dynamics with a surrogate model and estimating its structure via an alternating direction optimization technique. The estimated parameters are further refined using a forward solver–based estimation algorithm, and the time labels are recovered by projecting observation particles onto the simulated solution curve. Our method rapidly and robustly produces an admissible initial guess for the underlying system and its parameters, enabling accurate and efficient reconstruction of both the hidden system and time labels. Furthermore, our approach is applicable to unlabeled data with arbitrary observation time distributions and demonstrates robust performance in the presence of noise.
    

The organization of the paper is as follows: In Section 2, we give the mathematical formulation of the problem. Subsequently, Section 3 explains the process in which our method estimates the parameter of the hidden dynamics and reconstructs the time label, wherein we partition the algorithm into two distinct phases. Experimental results and some discussions are summarized in Section 4, followed by the conclusion of our research and an exploration of prospective avenues in Section 5.

%% file: problem.tex
\section{Problem Statement}\label{problem}
Consider a general scenario in which we collect data from $L$ distinct trajectories of an unknown dynamical system, each initiated from a different initial condition.  For the $l$-th trajectory, observation instants  are modeled as random variables $t^l\sim\nu_{t^l}$ with support on $[t_0^l,t_0^l+T^l]$(e.g., following a uniform distribution $U(t_0^l, t_0^l + T^l)$) rather than as deterministic points. Here, we assume that the underlying dynamical system governing each trajectory 
$\tmmathbf{x}^l (t)$ evolves according to an autonomous ODE, with each trajectory initiated from a distinct initial condition
\begin{equation}
  \left\{ \begin{array}{c}
    \frac{d\tmmathbf{x}^l}{\tmop{dt}} =\tmmathbf{f} (\tmmathbf{x}^l,
    \tmmathbf{\theta}^{\star}), t \in [t_0^l, t_0^l + T^l]\\
    \tmmathbf{x}^l (t_0^l) = (\tmmathbf{x}^l_1 (t_0), \ldots, \tmmathbf{x}_d
    (t_0))^{\top}\in\mathbb{R}^d .
  \end{array} \right. \label{ODE}
\end{equation}
Therefore, the observation samples $\boldsymbol{x}^l(t^l)\sim\rho_{\boldsymbol{x}^l} = \tmmathbf{x}^l_{\#}\nu_{t^l}$ is a random vector whose distribution $\rho_{\boldsymbol{x}^l}$ is obtained by push-forwarding $\nu_{t^l}$ through $\tmmathbf{x}^l(\cdot)$. In practice, to account for measurement noise $\delta$ (e.g., white noise), we denote the noisy data distribution as $\rho^{\delta}_{\boldsymbol{x}^l}$.

The task is now to find the best forcing term $\tmmathbf{f} (\tmmathbf{x},
\tilde{\tmmathbf{\theta}})$ such that all $L$ random vectors obtained by the resulting dynamical system $\boldsymbol{x}^l_{\theta}(t^l)\sim\rho_{\boldsymbol{x}_{\theta}^l} = \tmmathbf{x}_{\theta\#}^l\nu_{t^l}$ the corresponding  observation $\rho^{\delta}_{\boldsymbol{x}^l}$ in the sense of distribution. To this end, we formulate our goal into a minimization problem
\begin{equation}
  \begin{array}{lc}
\mathop{\min}\limits_{\tmmathbf{\theta}\in\Theta}& \sum_{l=1}^L \mathcal{D}(\tmmathbf{x}_{\theta\#}^l\nu_{t^l}, \rho^{\delta}_{\boldsymbol{x}^l}) \\
  s.t. & \left\{\begin{array}{ccl}
  \frac{d\tmmathbf{x}_{\tmmathbf{\theta}}^l}{dt}(t) &=& \tmmathbf{f} (\tmmathbf{x}_{\tmmathbf{\theta}}^l,
    \tmmathbf{\theta}), t \in [t_0^l, t_0^l + T^l] \\
  \tmmathbf{x}_{\tmmathbf{\theta}}^l(t_0^l) &=&   \tmmathbf{x}^l(t_0^l)
  \end{array}. \right.
   \end{array} \label{OriginProblem}
 \end{equation}
 Here, $\Theta$ denotes the parameter space to which $\tmmathbf{\theta}$ belongs. Here we assume that the initial condition of each trajectory $\tmmathbf{x}^l(t_0^l)$ is known. The formulation \eqref{OriginProblem} defines an inverse data-matching problem, in which $\mathcal{D}$ denotes a metric or divergence on the space of probability measures. Many classes of discrepancy measures can be employed in this context; in the present work, we focus on a widely used metric—the quadratic Wasserstein metric. Specifically, the quadratic Wasserstein distance between two probability measures is defined as:
\[
\mathcal{W}_{2}(\mu, \nu)=\left(\min _{\gamma \in \Gamma(\mu, \nu)} \int\|x-y\|^{p} \mathrm{~d} \gamma\right)^{\frac{1}{2}}
\]
where $\Gamma$ represent the set of all couplings between the two probability measures. In practical scenarios, the discrete-time observed data of the \(l\)-th trajectory is denoted by \(\mathbb{X}^l = \{\mathbf{x}^l_1, \mathbf{x}^l_2, \ldots, \mathbf{x}^l_n\}\); these data are generated from process (\ref{ODE}) at \(n\) time points \(\mathbf{t}^l = \{t^l_1, \ldots, t^l_n\}\), sampled i.i.d. from \(\nu_{t^l}\). Numerous studies have proposed methods to estimate the Wasserstein distance directly from the observed samples—without explicitly estimating the underlying distribution function—which is a key motivation for our focus on the Wasserstein distance\cite{kolouri2019generalized,nguyen2022hierarchical,nguyen2024energy}. Furthermore, another primary objective of our work is to reconstruct the unknown time labels \(\mathbf{t}^l_i\) for each data point \(\mathbf{x}^l_i\).

.

%% file: method.tex
\section{Methodology}
In this section, we present a two-phase learning framework for extracting the underlying dynamical system from data lacking time labels.

By leveraging the sliced Wasserstein distance (SWD) and dictionary representation, we formalize the problem as an optimization task and propose a forward solver-based optimization algorithm. This algorithm solves a parameterized ODE system at a batch of time instants sampled from the observation time distribution, and then minimizes the SWD between the generated trajectories and the observed samples via gradient-based optimization. We refer to this process as the parameter identification phase. However, for complex systems, this method requires a good initial guess to converge stably to the correct form. Inspired by the frameworks presented in \cite{yangGenerativeEnsembleRegression2021} and \cite{chenPhysicsinformedLearningGoverning2021}, we propose a deep learning–based approach that substantially accelerates the initial training phase and yields a more accurate estimation for subsequent forward solver-based correction. We denote this subsequent stage as the distribution matching phase. In this phase, a neural network approximates the dynamical system's solution by iteratively minimizing the SWD, while physics-informed regularization is concurrently applied to enforce the requisite smoothness of the solution function and to provide an estimate of the hidden dynamics.

Since computational experiments reveal that short trajectories are easier to identify than long, intricate trajectories, we propose to initially partition the long trajectory into short segments using an unsupervised clustering technique for systems exhibiting complex phase spaces.
\subsection{Parameter Identification}\label{sec:pi}
We now investigate a forward solver-based approach for solving problem \eqref{OriginProblem}\ by adapting the four steps of FSNLS to the setting of unlabeled data. We assume that the physical law is governed by only a few essential terms chosen from a large library of candidate functions; accordingly, we adopt dictionary representations for the unknown terms in $\tmmathbf{f}(\tmmathbf{x},\tmmathbf{\theta})$. Let 
\[ \tmmathbf{\phi} (\tmmathbf{x}) = \left[
     \tmmathbf{f}_1 (\tmmathbf{x}),  \tmmathbf{f}_2 (\tmmathbf{x}),  \ldots
      ,\tmmathbf{f}_s (\tmmathbf{x})
   \right] . \]
denotes the dictionary with $s$ candidate functions $\tmmathbf{f}_s:\mathbb{R}^{d}\rightarrow \mathbb{R}^{d}$, then the forcing term can be represented as the linear combination of all candidates with weight $\tilde{\tmmathbf{\theta}} \in\mathbb{R}^s$.

Comparing the discrepancy between two large-scale, high-dimensional point clouds using the Wasserstein distance is computationally challenging; therefore, we pursue an approach that directly estimates the Wasserstein distance from samples. By leveraging the closed-form solution of the Wasserstein distance in one dimension, the Sliced Wasserstein distance (SWD) has been demonstrated to be an efficient metric for comparing probability distributions, particularly in high-dimensional settings.  Formally, Sliced Wasserstein distances consider the average or maximum of Wasserstein distances between
one-dimensional projections of the two distributions\cite{nietert2022statistical}
\begin{equation}\label{eq:swd}
\underline{\mathrm{SW}}_{2}(\mu, \nu):=\left[\int_{\mathbb{S}^{d-1}} \mathrm{~W}_{2}^{2}\left(\mathfrak{p}_{\sharp}^{\omega} \mu, \mathfrak{p}_{\sharp}^{\omega} \nu\right) d \sigma(\omega)\right]^{1 / 2} \text { and } \quad \overline{\mathrm{SW}}_{2}(\mu, \nu):=\max _{\omega \in \mathbb{S}^{d-1}} \mathrm{~W}_{2}\left(\mathfrak{p}_{\sharp}^{\omega} \mu, \mathfrak{p}_{\sharp}^{\omega} \nu\right),
\end{equation}
where where $\mathfrak{p}_{\sharp}^{\omega} \mu$ is the pushforward of $\mu$ under the projection $\mathfrak{p}^{\omega}:x\rightarrow \omega^{\top}x$ from $\mathbb{R}^d$ to $\mathbb{R}$ and $\sigma$ is the
uniform distribution on the unit sphere $\mathbb{S}^{d-1}$. Leveraging the explicit solution of 1D Wasserstein distance, the 
Sliced Wasserstein distances between empirical distributions can be approximated via order statistic. Indeed, let $
\hat{\mu}_n := n^{-1} \sum_{i=1}^n \delta_{X_i}$ and $\hat{\nu}_n := n^{-1} \sum_{i=1}^n \delta_{Y_i}
$ be the empirical distributions of samples $X_1, \ldots, X_n$ and $Y_1, \ldots, Y_n$. For any given direction \(\omega \in \mathbb{S}^{d-1}\), let \(X_i(\omega) = \omega^\top X_i\) denote the projected samples. By Lemma 4.2 in [8], the order statistics of these projections, namely \(X_{(1)}(\omega) \leq \cdots \leq X_{(n)}(\omega)\) and \(Y_{(1)}(\omega) \leq \cdots \leq Y_{(n)}(\omega)\), yield the one-dimensional Wasserstein distance
\begin{equation}
W_2^2(\mathfrak{p}_{\sharp}^\omega\hat{\mu}_n, \mathfrak{p}^\omega_{\sharp}\hat{\nu}_n) = \frac{1}{n} \sum_{i=1}^n \left| X_{(i)}(\omega) - Y_{(i)}(\omega) \right|^2.
\end{equation}
The sliced distances \(\underline{SW}_2\) and \(\overline{W}_2\) are then approximated by integrating or taking the maximum of the above expression over \(\omega \in \mathbb{S}^{d-1}\)

.\begin{equation}\label{eq:swda}
\underline{\mathrm{SW}}_{2}^2(\hat{\mu}_n, \hat{\nu}_n)\approx \frac{1}{nm}\sum_{j=1}^m  \sum_{i=1}^n \left| X_{(i,j)}(\omega_j) - Y_{(i,j)}(\omega_j) \right|^2, \end{equation}\begin{equation}\label{eq:swdm}\overline{\mathrm{SW}}_{2}^2(\hat{\mu}_n, \hat{\nu}_n)\approx\max _{\omega_j^{\star} \in \{\omega_j\}_{j=1}^m}  \frac{1}{n}\sum_{i=1}^n \left| X_{(i,j)}(\omega_j) - Y_{(i,j)}(\omega_j) \right|^2.
\end{equation}
where $\{\omega_j\}_{j=1}^m$ is sampled from uniform distribution on $\mathbb{S}^d$ and $X_{(i,j)}(\omega_j)$ is the orther statistics for $\{X_i(\omega_j)\}_{i=1}^n$.
In this paper, we adopt \(\underline{\mathrm{SW}}_2\) as the discrepancy metric and compute it using pyOT \cite{flamary2021pot}, an open-source Python library. For convenience, we denote \(\underline{\mathrm{SW}}_2\) for empirical distributions by \(\mathrm{SW}_2\). 

In the field of system identification, researchers typically aim to extract models from data that are both accurate and simple. The principle of sparsity in modeling is fundamentally rooted in Occam’s razor\cite{domingos1999role}, which posits that in explaining the latent dynamics within data, the primary goal should be to select the simplest possible model. By favoring models with the fewest nonzero coefficients, one not only enhances interpretability but also mitigates the risk of overfitting. To this end, we introduce an 
$l_0$ penalty as the regularization term.Together with the sliced Wasserstein distance and dictionary representation, the computational optimization problem is reformulated as follows:
 \begin{equation}
  \begin{array}{ll}
\mathop{\min}\limits_{\tmmathbf{\theta}\in\Theta}& \sum_{l=1}^L SW_2^2(\tmmathbf{x}_{\theta\#}^l\nu_{t^l}, \rho^{\delta}_{\boldsymbol{x}^l}) +  \lambda_{\text{0}}\|\tmmathbf{{\theta}}\|_{0}\\
  s.t. & \left\{\begin{array}{ccl}
  \frac{d\tmmathbf{x}_{\tmmathbf{\theta}}^l}{dt}(t) &=& \tmmathbf{\phi} (\tmmathbf{x}^l_{\tmmathbf{\theta}})\tmmathbf{\theta}, t \in [t_0^l, t_0^l + T^l] \\
  \tmmathbf{x}_{\tmmathbf{\theta}}^l(t_0^l) &=&   \tmmathbf{x}^l(t_0^l)
  \end{array}. \right.
   \end{array} \label{ComputeProblem}
 \end{equation}
Here, \(\rho^{\delta}_{\boldsymbol{x}^l}\) and \(\tmmathbf{x}_{\theta\#}^l\nu_{t^l}\) represent empirical distributions obtained via sampling. To sample \(\tmmathbf{x}_{\theta\#}^l\nu_{t^l}\), we approximate the target function by discretizing the approximated ODE at sampled time instants corresponding to the current estimate of \(\tmmathbf{\theta}\). Specifically, for the \(l\)-th trajectory, we first sample \(\{t_i^l\}_{i=1}^B\) i.i.d. from \(\nu_{t^l}\) and uniformly select \(\{\tmmathbf{x}_{ob,i}^l\}_{i=1}^B\) from the observation data $\mathbb{X}_l$. Subsequently, samples of \(\tmmathbf{x}_{\theta\#}^l\nu_{t^l}\) are obtained by solving the ODE at these designated time instants.
\begin{equation}
    \tmmathbf{x}_{\tmmathbf{\theta}}^l (t_i^l)=\textit{ODESolve}\left(\tmmathbf{\phi} (\tmmathbf{x}^l_{\tmmathbf{\theta}})\tmmathbf{\theta},\tmmathbf{x}(t_0^l),[t_0^l,t_{i}^l]\right).\label{SampleODE}
\end{equation}
Here, the batch size \(B\) and the number of projection directions \(m\) must be chosen sufficiently large to accurately approximate the distributions \(\tmmathbf{x}_{\theta\#}^l\nu_{t^l}\) and \(\rho^{\delta}_{\boldsymbol{x}^l}\), as well as to ensure reliable Monte Carlo integration in the sliced Wasserstein distance. For multi-trajectory observation, the sampling process of each trajectory can be parallelly computed, and the distributional loss of each piece by \eqref{eq:swda}
\begin{equation}\label{eq:loss_l}
  L_{l} = SW_2^2(\hat{\rho}_{\{ \tmmathbf{x}_{\tmmathbf{\theta}}^l (t_i^l)\}_{i=1}^B},\hat{\rho}_{\{\tmmathbf{x}_{ob,i}^l\}_{i=1}^B} )=  \frac{1}{mn}\sum_{j=1}^m  \sum_{i=1}^B \left| (\tmmathbf{x}_{\tmmathbf{\theta}}^l (t_i^l))_{(i,j)}(\omega_j) - (\tmmathbf{x}_{ob,i}^l) _{(i,j)}(\omega_j) \right|^2,
\end{equation}
where \(\hat{\rho}\) denotes the empirical distribution of the given samples, and \(\tmmathbf{x}_{\tmmathbf{\theta}}^l(t_i^l)_{(i,j)}(\omega_j)\) represents the order statistics of \(\{\tmmathbf{x}_{\tmmathbf{\theta}}^l(t_i^l)(\omega_j)\}_{i=1}^B\), with the remaining terms defined analogously.

Due to the presence of the \(l_0\) penalty, the optimization task is highly intractable. Here, we employ proximal gradient descent to address this problem. Specifically, let \(\alpha\) denote the step size for the smooth component; then the proximal gradient descent update rule is given by
\begin{equation}
\tmmathbf{\theta}^{k+1} = \operatorname{prox}_{\alpha\lambda_0 \|\cdot\|_{0}}\Bigl(\tmmathbf{\theta}^{k}  - \alpha \nabla_{\theta}SW_2^2(\hat{\rho}_{\{ \tmmathbf{x}_{\tmmathbf{\theta}^k}^l (t_i^l)\}_{i=1}^B},\hat{\rho}_{\{\tmmathbf{x}_{ob,i}^l\}_{i=1}^B} ) \Bigr).
\end{equation}
where the proximal operator of the \(l_0\) regularization term is given by the hard thresholding operator applied to each element \(\theta_i\).
\begin{equation}\label{proximal}
\operatorname{prox}_{\alpha\lambda\|\cdot\|_0}(\tmmathbf{\theta}_{i}) = H_{\sqrt{2\alpha\lambda}}(\tmmathbf{\theta}_{i}) =
\begin{cases}
\tmmathbf{\theta}_{i}, & \text{if } |\theta| > \sqrt{2\alpha\lambda}, \\
0,   & \text{otherwise.}
\end{cases}
\end{equation}
Algorithm \ref{alg:PI} summarizes the procedure for the parameter identification phase, wherein we initially disregard the regularization term and perform a warm-up optimization to capture the system's behavior.

\begin{algorithm}
\caption{Parameter identification phase algorithm}
\label{alg:PI}
\begin{algorithmic}
\STATE{\textbf{Input}: $ \{\mathbb{X}_l\}_{l=1}^L,\{\tmmathbf{x}^l(t_0^l)\}_{l=1}^{L},\{\nu_{t^l}\}_{l=1}^L,Iter,B, M,\lambda_0, \alpha,\text{Iter}_{warm}$}
\FOR{$\text{it}=1:\text{Iter}$}
\STATE{$L_{\text{total}} = 0$}
\FOR{$l=1:L$}
\STATE{Sample $\{t_i^l\}_{i=1}^B \sim \nu_{t^l}$}
\STATE{Uniformly sample $ \{ \tmmathbf{x}_{ob,i}^l \}_{i = 1}^B$ from $\mathbb{X}_l$}
\STATE{Obtain $\{\tmmathbf{x}^l_{\tmmathbf{\theta}_k} (t_i^l)\}_{i=1}^B$ by \eqref{SampleODE}}
\STATE{Compute SWD loss $L_l$ by \eqref{eq:loss_l}, $L_{\text{total}} = L_{\text{total}} + L_l$ }
\ENDFOR
\STATE{$\tmmathbf{\theta_{k+1}} = \tmmathbf{\theta_k} - \alpha \nabla_{\theta} L_{\text{total}}(\tmmathbf{\theta_{k}})$ 
}
\IF{$k>\text{Iter}_{\text{warm}}$}
\STATE{$\tmmathbf{\theta}^{k+1}_j = \operatorname{prox}_{\alpha\lambda\|\cdot\|_0}(\tmmathbf{\theta}^{k+1}_j)$}
\ENDIF
\ENDFOR
\RETURN $\tmmathbf{\theta}_{\text{Iter}}$
\end{algorithmic}
\end{algorithm}

\begin{figure}[!htbp] 
    \centering
    \subfigure[]{\includegraphics[width=0.3\textwidth]{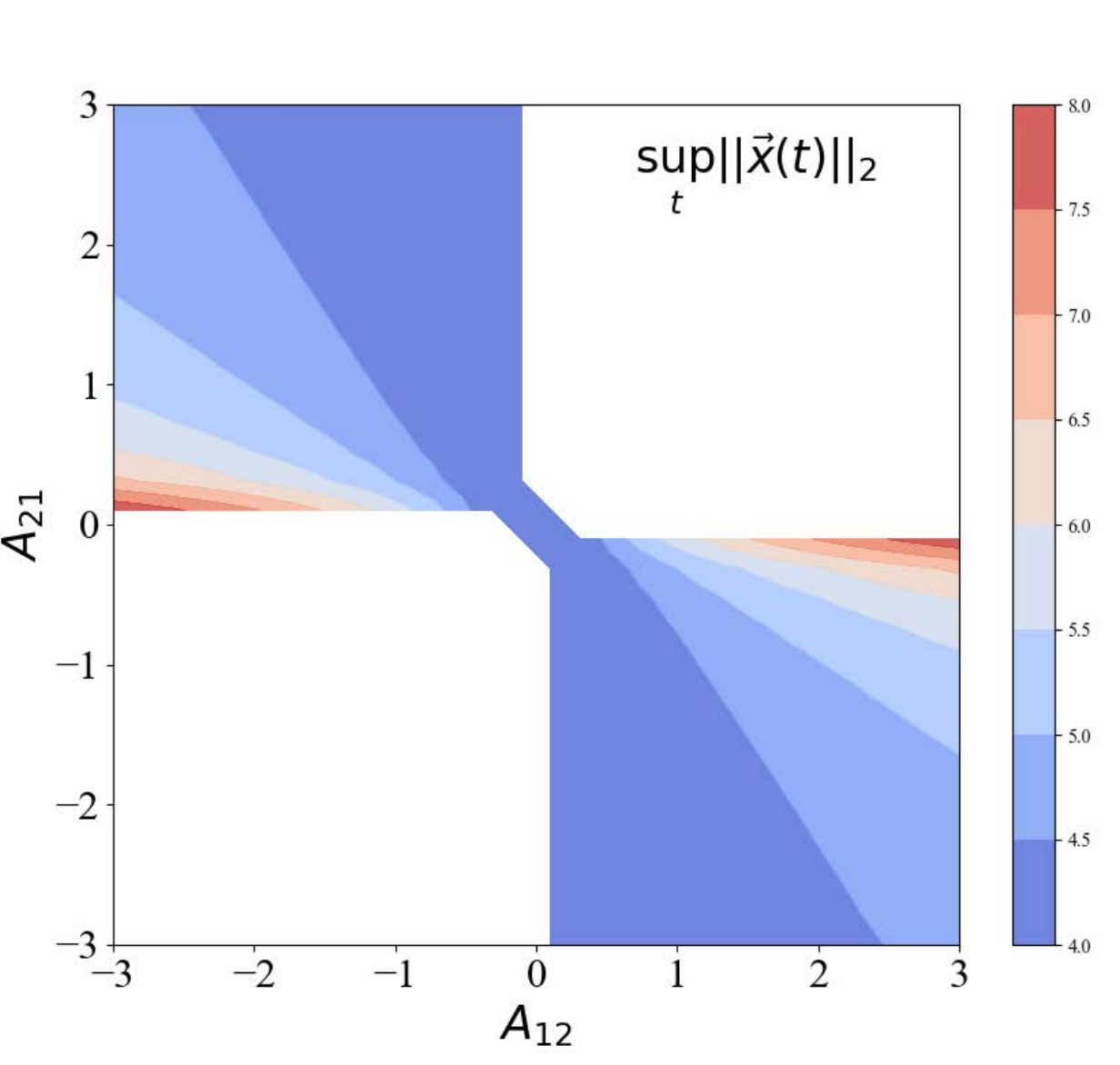}}
    \hspace{0.05\textwidth}
    \subfigure[]{\includegraphics[width=0.3\textwidth]{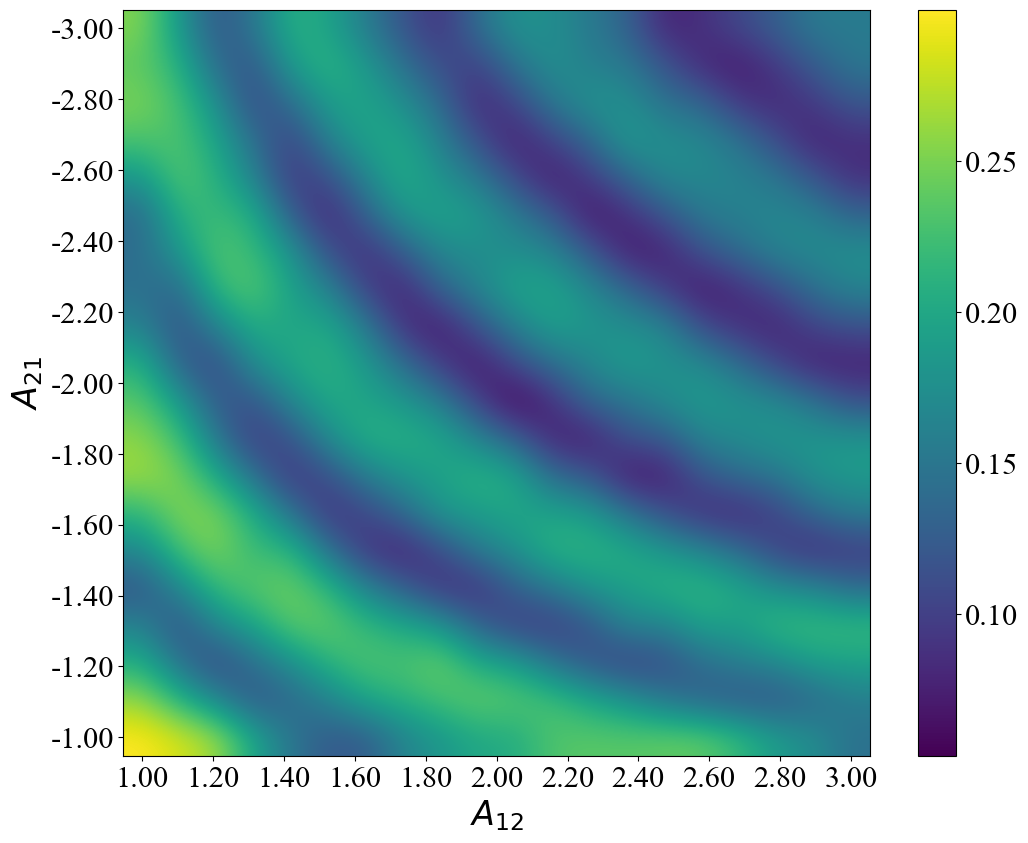}}\\[0.5ex]
    \subfigure[]{\includegraphics[width=0.3\textwidth]{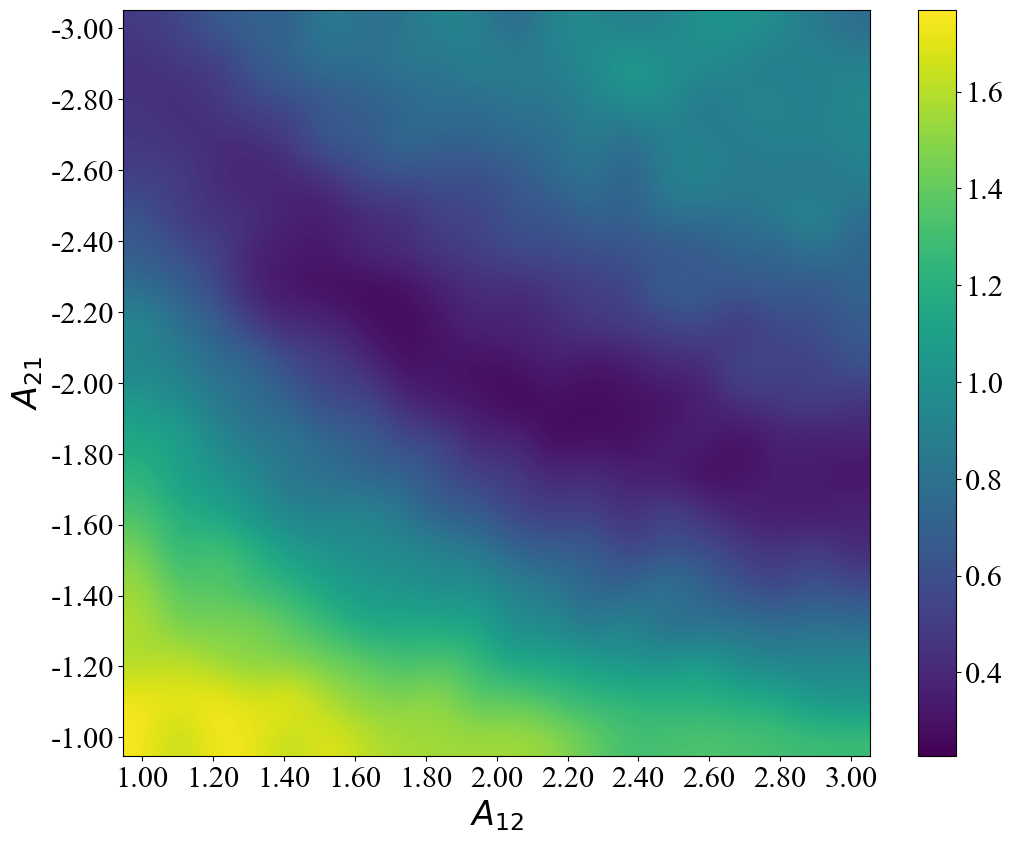}}
    \hspace{0.05\textwidth}
    \subfigure[]{\includegraphics[width=0.3\textwidth]{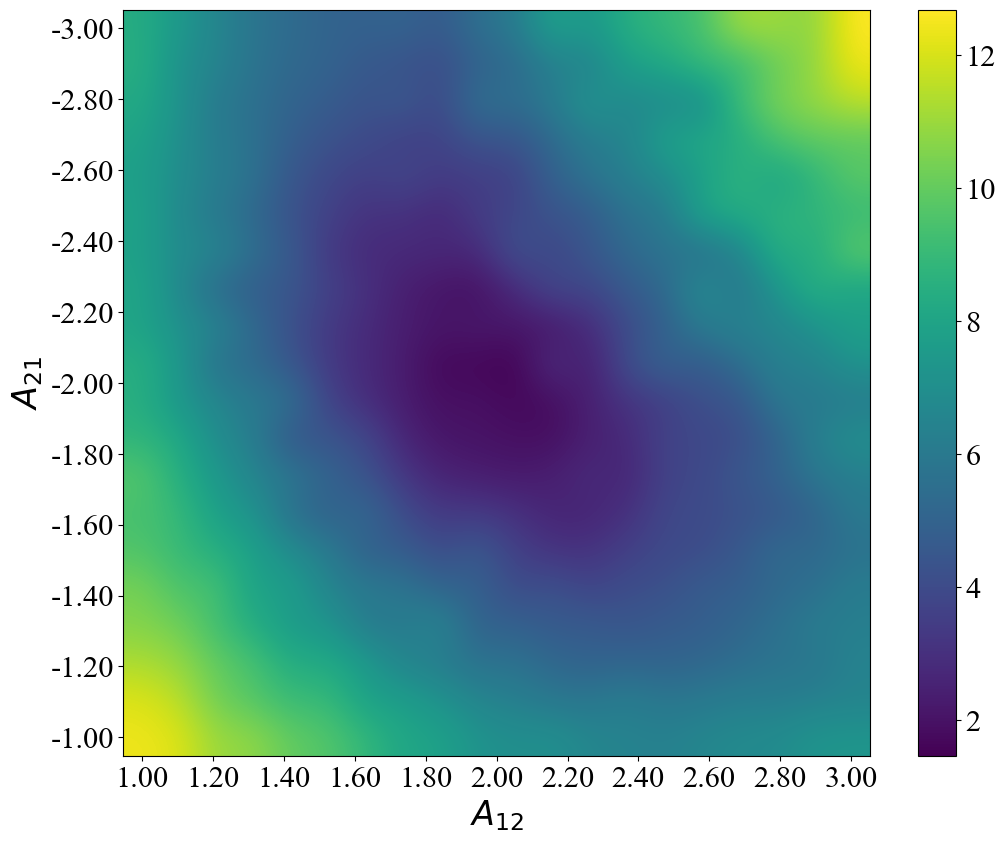}}
    \caption{\textit{Loss landscape analysis of Cubilc2D problem:} (a) The estimated $\sup_{t\in[0,10]}||\boldsymbol{x_{t}}||_2$ of parameter $(A_{12},A_{21})$ choice in $[1,3]\times[-3,-1]$; (b) The SWD loss landscape of $A_{12},A_{21}$ in $[1,3]\times[-3,-1]$ with $T=10$; (c) The SWD loss landscape of $A_{12},A_{21}$ in $[1,3]\times[-3,-1]$ with $T=0.4$; (d) ...}
    \label{loss_landscape}   
\end{figure}

Leveraging the identified parameters $\tmmathbf{\theta}$ we recovery the observation time of each data point by solving the
IVP at a dense uniform grid $\{ \hat{t}_i^l \}_{i = 1}^{M}\subset [t_0^l,t_0^l+T_l]$ for trajectory $l$ ($M$ is sufficiently large). The unlabeled observation $\mathbb{X}_l$ are then projected onto the obtained labeled trajectory $ \tmmathbf{x}_{\tmmathbf{\theta}} (\hat{t}_i^l)$ by solving an optimization problem
\begin{equation}
  \min_{\{t_j^l \}_{j = 1}^n \subset \{ \hat{t}^l_i \}_{i = 1}^M}  \sum_{j = 1}^n
  \|\tmmathbf{x}_j^l -\tmmathbf{x}^l_{\tmmathbf{\theta}} (t_j^l)\|_2^2 .
\end{equation}
We observe that the practical application of the forward solver‐based optimization method is hindered by several limitations. Figure~\ref{loss_landscape} shows some results given by above algorithm for an illustrative example.




\[\left\{
    \begin{array}{ll}
  \frac{dx_1}{dt} = -0.1x_1^3+2x_2^3     \\
 \frac{dx_2}{dt} = -2x_1^3-0.1x_2^3      \\
 \boldsymbol{x}(0) = (4,0)^\top.
\end{array}\right.\]

Figure~\ref{loss_landscape}(a) presents a contour map of \(\sup_{t\in[0,10]} \|\boldsymbol{x}_t\|_2\) over the parameter space \(\{(A_{12},A_{21}) \mid (A_{12},A_{21}) \in [1,3] \times [-3,-1]\}\), revealing that initializing parameters in the first or third quadrant yields blow-up solutions and precludes further updates. ($A_{12}$ and $A_{21}$ represents the coefficient of $x_2^3 $ and $x_{1}^3$) This result show that a good initial guess is important in the algorithm, otherwise the ODE solver may blowup. Then it is impossible to get a good reconstruction for the underlying dynamic system. 

Figure~\ref{loss_landscape}(b) illustrates that the loss function exhibits marked non-convexity, such that solutions initialized near the optimum tend to become trapped in local minima. However, this non-convexity is substantially alleviated for shorter trajectories; indeed, Figure~\ref{loss_landscape}(c) reveals an almost convex loss function in the same parameter region when observations are limited to instances \(\boldsymbol{t}\in[0,0.4]\). Moreover, Figure~\ref{loss_landscape}(d) further demonstrates that the loss conditions defined as the average SWD across multiple  trajectories are significantly improved when the complex trajectory observations are segmented into ten shorter trajectories by Algorithm \ref{alg:TS}. These findings suggest that partitioning complex trajectory observations into shorter segments may reduce overall complexity.

Based on the results in Figure~\ref{loss_landscape}, we will introduce methods for initialization and trajectory segmentation in subsection \ref{initialization} and \ref{segment} respectively. 

\subsection{Initialization by Distribution Matching}
\label{initialization}
To address the aforementioned problem, we introduce a novel method that leverages a deep neural network surrogate model for solving the ODE \(\tmmathbf{x}_{\tmmathbf{\theta}}^l(t)\) to obtain an admissible estimate of \(\tmmathbf{\theta}\). In this approach, the DNN is trained to approximate the transport map from the observation time distribution to the estimated data distribution.

Specifically, we employ \(L\) distinct deep neural networks, denoted by \(\tmmathbf{x}_{\psi_l}^l(\cdot)\), to approximate the solution functions for the \(L\) observation trajectories, given that their initial conditions differ and the observation times may overlap. By introducing a residual physics loss $\tmop{L}_{\text{Phy}}$ as well as a soft constraint on the initial condition, the original problem \eqref{OriginProblem} is modified as
\begin{equation}\label{eq:dm_target1}
\min_{\psi_l\in \Psi,\boldsymbol{\theta}\in \Theta}\sum_{l=1}^L \left[
\underbrace{SW_2^2\Bigl(\tmmathbf{x}_{\psi_l\#}^l\nu_{t^l},\, \rho^{\delta}_{\boldsymbol{x}^l}\Bigr)}_{\text{Data loss}}
+ \underbrace{\lambda_{\text{init}} \Bigl\|\tmmathbf{x}_{\psi_l}^l(t_0^l)-\tmmathbf{x}^l(t_0^l)\Bigr\|_2^2}_{\text{IC loss}}
+ \underbrace{\lambda_{\tmop{Phy}}\,\tmop{L}_{\text{Phy}}\Bigl(\tmmathbf{x}_{\psi_l}^l, \boldsymbol{\theta}\Bigr)}_{\text{Physics loss}}
\right].
\end{equation}

where $\Psi$ is the parameter space of the neural networks surrogate and $\lambda_{\text{init}},\lambda_{\text{Phy}}$ is the weight of the initial loss and the Physics loss. We can compute the \textit{Data loss} term using an approach analogous to that in \eqref{SampleODE} and \eqref{eq:loss_l}. In particular, we sample \(\{t_i^l\}_{i=1}^B\) and \(\{\tmmathbf{x}_{ob,i}^l\}_{i=1}^B\) in the same manner, and directly evaluate the solution \(\{\tmmathbf{x}_{\psi}(t_i^l)\}_{i=1}^B\) at each time instant to obtain samples from \(\tmmathbf{x}_{\psi_l\#}^l\nu_{t^l}\). the \textit{Data loss} is calculated by the Monte Carlo approximation of SWD
\begin{equation}\label{eq:loss_dm_data}
  SW_2^2(\hat{\rho}_{\{ \tmmathbf{x}_{\psi_l}^l (t_i^l)\}_{i=1}^B},\hat{\rho}_{\{\tmmathbf{x}_{ob,i}^l\}_{i=1}^B} )=  \frac{1}{mn}\sum_{j=1}^m  \sum_{i=1}^n \left| (\tmmathbf{x}_{\psi_l}^l (t_i^l))_{(i,j)}(\omega_j) - (\tmmathbf{x}_{ob,i}^l) _{(i,j)}(\omega_j) \right|^2
\end{equation}

Once the generated distribution $\tmmathbf{x}_{\psi_l\#}^l\nu_{t^l}$ approximate the data distribution $\rho^{\delta}_{\boldsymbol{x}^l}$ in discrete sense, one can extract the underlying dynamics with traditional system identification method like FSNLS or SINDy from deterministic functions $\tmmathbf{x}_{\psi_l}$.However, regression in the distributional sense does not constrain the network’s regularity, and the resulting surrogates may be challenging to distill due to insufficient smoothness. Inspired by \cite{chenPhysicsinformedLearningGoverning2021}, we proposed introducing a Physics-informed loss term to unify the tasks of approximating the neural surrogate and extracting the hidden dynamics, thereby addressing this issue.  In the continuous case, the Physics-informed loss is formulated as the norm of the residual derived from the reconstructed ODE
\begin{equation}\label{eq:PI_reg}\tmop{Reg}_{\tmmathbf{\theta}} (\tmmathbf{x}_{\psi_l}) = 
\int_{t_0^l}^{t_0^l+T^l}\| \tmmathbf{\phi}
    (\tmmathbf{x}^l_{\psi_l}(t)) \tmmathbf{\theta}- \dot{\tmmathbf{x}}^l_{\psi_l}(t)\|_2^2 dt + \lambda_{\text{sparse}}\|\tmmathbf{\theta}\|_0, \end{equation}
   where the time derivative \(\dot{\tmmathbf{x}^l_{\psi}}(t)\) is computed by automatic differentiation, and the \(\ell_0\) penalty is similarly incorporated to account for the system's sparsity. We hope to simultaneously adjust the deep neural network parameters \(\psi_l\) and the ODE coefficients \(\boldsymbol{\theta}\) so that the network not only fits the data but also satisfies the constraints imposed by the underlying ODE, but this transforms Problem \eqref{eq:dm_target1} into a joint optimization task both the neural network parameters \(\psi_l\) and \(\boldsymbol{\theta}\), while the \(l_0\) penalty further complicates the optimization process. To account for this, we propose to split the problem \eqref{eq:dm_target1} into two trackable subproblems and perform Alternating Direction Optimization(ADO) between the subproblems. Indeed, our algorithm iteratively alternates between fixing \(\tmmathbf{\theta}\) to optimize \(\psi_l\) and fixing \(\psi_l\) to refine \(\tmmathbf{\theta}\) for the problem \eqref{eq:dm_target1}. This approach resembles several well-established optimization techniques.In particular, it is closely related to the so-called block coordinate descent method since we alternatingly update $\psi_l$ and $\tmmathbf{\theta}$, the coordinate blocks in our problem. \\
   \textbf{From $\psi_l$ to $\tmmathbf{\theta}$:} Here, we fix the estimated solution function \(\tmmathbf{x}_{\psi}\) and employ the rectangle rule for the integration in \eqref{eq:PI_reg}. The resulting minimization objective for \(\tmmathbf{\theta}\) is equivalent to a standard least-squares problem with an \(l_0\) penalty
\begin{equation} \label{eq:regression}\min_{\tmmathbf{\tmmathbf{\theta}}} \| \tmmathbf{\phi} (\tmmathbf{X}_{\psi})
  \tmmathbf{\theta}- \dot{\tmmathbf{X}_{\psi}} \|_2^2 + \lambda_{\tmop{sparse}} \|
   \tilde{\tmmathbf{\theta}} \|_0. \end{equation}
where we evaluate the approximate solutions on uniform time grids and concatenate them in $\tmmathbf{X}_{\psi} = [\tmmathbf{x}_{\psi_1}^1(t_0^1),\ldots,\tmmathbf{x}_{\psi_1}^1(t_0^1+T^1),\ldots,\tmmathbf{x}_{\psi_L}^L(t_0^L),\ldots, \tmmathbf{x}_{\psi_L}^L(t_0^L+T^l)]^\top$, and $\dot{\tmmathbf{X}_{\psi}}$ denotes the time derivative of $\tmmathbf{X}_{\psi}$ evaluated by auto-grad. Problem \eqref{eq:regression} is a common problem in the field of system identification, and sequential threshold ridge regression (STRidge) is a widely used and effective algorithm for solving it. By iteratively applying a threshold to set small coefficients to zero after each ridge regression step, STRidge identifies the most significant factors within the candidate library, thereby yielding sparse models. \\
\textbf{From $\tmmathbf{\theta}$ to $\psi_l$:} For a given \(\tmmathbf{\theta}\), every term in the optimization problem \eqref{eq:dm_target1} is computable, and the \(l_0\) penalty term vanishes. Such problems are common in the study of physics-informed neural networks. Drawing on extensive research in PINNs, it is well established that gradient-based optimization methods can be employed to learn these problems.

We refer to this alternating direction optimization algorithm as the distribution matching phase, and present a summary of the procedure in Algorithm \ref{alg:DM}.
\begin{algorithm}
\caption{Distribution matching phase algorithm}
\label{alg:DM}
\begin{algorithmic}
\STATE{\textbf{Input}: $ \{\mathbb{X}_l\}_{l=1}^L,\{(t_0^l,T_l,\boldsymbol{x}^l(t_0^l))\}_{l=1}^{L},\{\nu_{t_l}\}_{l=1}^L,\text{Iter},\text{Iter}_{\text{warm}},\text{Iter}_{\text{update}},B$}
\FOR{$Iter=1:Iter$}
\FOR{$l=1:L$}
\STATE{Sample $\{t_i^l\}_{i=1}^{B} \sim \nu_{t_l}$, uniformly sample $\{ \tmmathbf{x}_{ob,i}^l \}_{i = 1}^B$ from $\tmmathbf{X}_l$}
\STATE{Compute $\{\boldsymbol{x}_{\psi_l}^l(t_i)\}_{i=1}^{B}$}
\IF{$\text{Iter}\leq$ $\text{Iter}_{\text{warm}}$}
\STATE{Update $\psi_l$ by only data loss and IC loss in \eqref{eq:dm_target1} }
\ELSE
\STATE{Compute total loss in \eqref{eq:dm_target1} and update $\psi_l$}
\ENDIF
\ENDFOR
\IF{$Iter \% Iter_{update}=0$}
\STATE{Update$\tmmathbf{\theta}$ by solving \eqref{eq:regression} with STRidge}
\ENDIF
\ENDFOR
\RETURN $\boldsymbol{x}_{\psi_l}^l,\boldsymbol{\theta}$
\end{algorithmic}
\end{algorithm}
By solving Problem \eqref{OriginProblem}, we can obtain a good estimate with reasonably accurate form and parameters via the distribution matching phase. Then we pass this result to parameter identification phase as initial guess. With this near-accurate estimate, only a few steps of the parameter identification phase are required to precisely identify the underlying system’s structure and parameters. Subsequent experiments demonstrate that the two-stage approach markedly enhances both accuracy and robustness compared to any single-stage methods; hence, we advocate this method.

\subsection{Trajectory Segmentation}
\label{segment}
In this subsection, we address the reconstruction problem using a single long trajectory. Our experiments (see Section~\ref{sec:pi}) demonstrate that simple, short trajectories exhibit a more favorable loss landscape than complex, long ones. Consequently, we propose to decompose the complex problem into several simpler subproblems through trajectory segmentation, thereby enhancing the solution capability of our approach.

Within the context of unlabeled data reconstruction, the trajectory segmentation problem encompasses two aspects: it involves not only grouping the data but also determining the initial condition and observation distribution for each cluster. For grouping the data, unsupervised clustering techniques—such as agglomerative clustering \cite{murtagh2014ward}, DBSCAN \cite{ester1996density}, and spectral clustering \cite{von2007tutorial}—can be employed to partition manifold-type data into a predetermined number of classes. To address the second aspect, we propose an estimation pipeline that functions as a post-processing step following the clustering procedure. Initially, we assume that the obtained trajectory segments (clusters) \(\{ \mathbb{X}_l \}_{l = 1}^L\) comprise data points drawn from distinct time intervals. To estimate the initial point and corresponding observation distribution, we designate as the primary trajectory the segment that contains the greatest number of elements within a neighborhood \(B_r(\tmmathbf{x}_0)\) of the initial point \(\tmmathbf{x}_0\). Thereafter, to determine subsequent trajectory pieces, we employ a set of rules to select the tail element of \(\mathbb{X}_l\) and the head element of the following segment \(\mathbb{X}_{l+1}\), beginning with the identified initial tail point \(\tmmathbf{x}_{\text{Tail},l}\).
$$\left\{\begin{array}{ccc}
 \tmmathbf{x}_{Tail,l} &= &\mathop{\arg\min}\limits_{\tmmathbf{x}\in\mathbb{X}_l
\backslash B_r({\tmmathbf{x}_{Head,l}})} \#\{\mathbf{y}|\mathbf{y}\in\mathbb{X}_l\cap B_r(\tmmathbf{x}) \}\\
\tmmathbf{x}_{\text{Head},l+1} &= &\mathop{\arg\max}\limits_{\tmmathbf{x}\in\mathbb{X}_{l+1}
} \#\{\tmmathbf{y}|\tmmathbf{y}\in\mathbb{X}_l\cap B_r(\tmmathbf{x}) \}
\end{array}\right. .
\label{Rule}
$$
where $\#$ denotes the element number of a set, and the next trajectory piece is selected as the one that intersects the most with $B_r(\tmmathbf{x}_{Tail,l})$.

\begin{figure}[htbp] 
	\subfigure[]{	\includegraphics[width=0.32\linewidth]{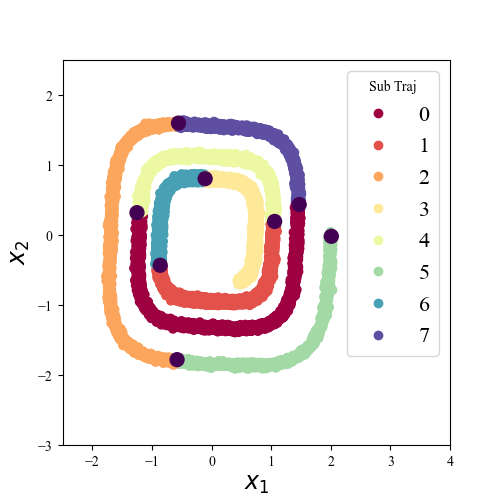}}
	\subfigure[]{
\includegraphics[width=0.32\linewidth]{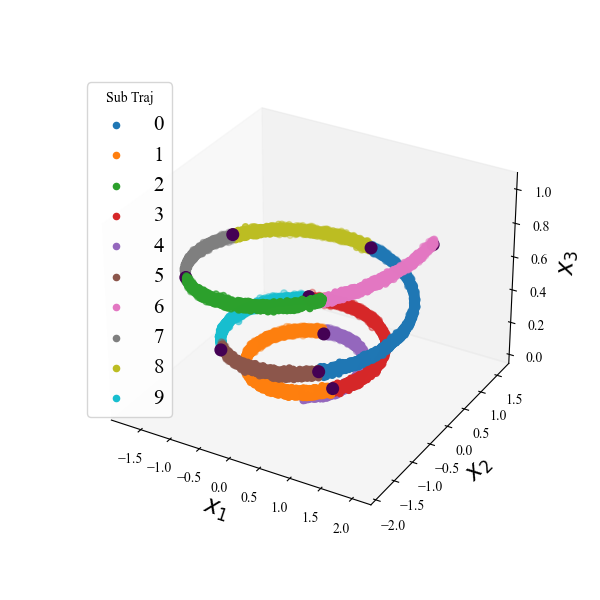}}
        \subfigure[]{		\includegraphics[width=0.32\linewidth]{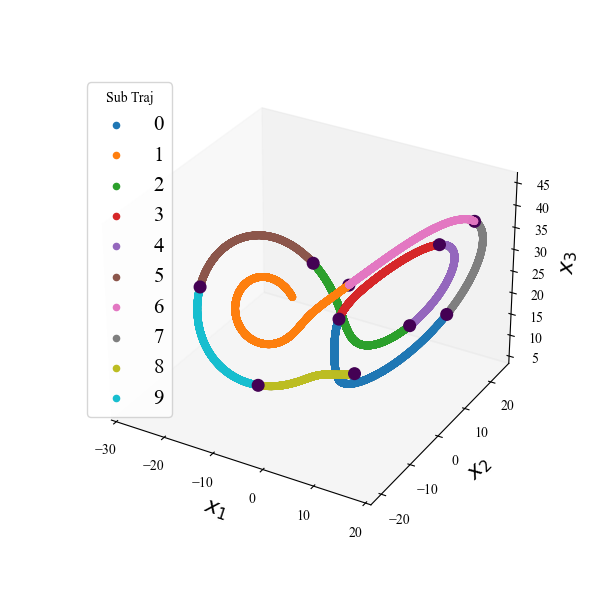}}
	\caption{\textbf{Segmentation result of illustrative problem:} (a) the 8 piece cluster result of Cubic2D ODE system initiate at $(2,0)$ with $T = 10$;(b) the 10 piece cluster result of Linear3D ODE system initiate at $(2,0,-1)$ with $T = 10$;(c) the 10 piece cluster result of Lorenz ODE system initiate at $(10,20,-10)$ with $T = 3$.}
	\vspace{0cm}
	\label{cluster}
\end{figure}

Finally, we employ a two-sided truncated distribution to estimate the observation distribution for each trajectory segment
\[t_l =t \mathbb{I}_{t_0^l <\tmmathbf{t}< P^{- 1}
   \left( P  (t_0^l) + \frac{n_l}{n}  \right)} \sim \nu_{t_l} \]
where \(n_l\) is the element number, \(t_0^l\) is the initial time, $t$ is the observation time of the original trajectory and $P$ is the probability distribution function of $\nu_{t}$. We use the frequency to approximate the proportional length of each trajectory segment. It should be noted that the segmentation process may occasionally fail due to overlapping and intertwined trajectory configurations—a matter we reserve for future investigation. The trajectory segmentation methodology is summarized in Algorithm \ref{alg:TS}, and the segmentation results for several systems are presented in Fig~\ref{cluster}.

\begin{algorithm}
\caption{Trajectory segmentation algorithm}
\label{alg:TS}
\begin{algorithmic}
\STATE{\textbf{Input}: $ \mathbb{X},\mathbb{P},\boldsymbol{x}(t_0),T,L$}
\STATE{$\{\mathbb{X}_l\}_{l=1}^L = \text{Agglomerative Clustering}(\mathbb{X},L)$}
\FOR{$l=1:L$}
\STATE{$t_0^{l+1} =P^{- 1}
   \left( P  (t_0^l) + \frac{n_l}{n}\right)$}
\STATE{Determine $\boldsymbol{x}(t_0^{l+1}) = \tmmathbf{x}_{\text{Head},l+1}$ by \eqref{Rule}}
\STATE{ $t_l =t \mathbb{I}_{t_0^l <\tmmathbf{t}< P^{- 1}
   \left( P  (t_0^l) + \frac{n_l}{n}  \right)} \sim \nu_{t_l}, T_l = t_0^{l+1}-t_0^l$}
\ENDFOR
\RETURN $\{(\mathbb{X}_l,t_0^l,T^l,\tmmathbf{x}(t_0^l),\nu_{t_l})\}_{l=1}^L$
\end{algorithmic}
\end{algorithm}
It is important to note that for the reconstruction problem of a single long trajectory, the segmented observation intervals do not overlap. Consequently, we employ a single neural network to approximate the solution functions for all segments concurrently, thereby leveraging the continuity inherent in neural networks.

%% file: experiment.tex
\section{Experiments}\label{sec:e}
In this section, we apply our methodology to the elementary demonstrative systems dpresented in \cite{bruntonDiscoveringGoverningEquations2016}, the Lorenz equations and Duffing equations in the chaotic regime, the Lotka-Volterra systems in high dimensional regime, the pendulum equations in non-polynomial regime. 
This study focuses on the challenging problem of reconstructing a single long trajectory, a task that inherently presents significant difficulties.  In the following, we assess the precision, robustness, and efficacy of our approach under uniform and non-uniform observation distributions, thereby underscoring its advantages.
\subsection{Numerical setups and performance metrics}
For data generation, in each experiment we first extract 50,000 samples from the observation time distribution to form a non-uniform time grid, and then use the LSODA method to solve the governing equations, thereby ensuring an equitable approximation of the distribution. For error computation, we obtain a reference numerical solution on a uniform time grid \(\{t_i^\star\}_{i=0}^{M}\) using the RK4 scheme. The estimated solution is obtained by employing the RK4 scheme on the uniform time grid \(\{t_i^\star\}_{i=0}^{M}\) to solve a series of subproblems corresponding to the estimated initial condition and time intervals. To evaluate performance, we compute the approximation error of the learned dynamics $\boldsymbol{E}_{\text{sol}}$, the error in the identified system parameters$\boldsymbol{E}_{\text{para}}$, and the error in the reconstructed time labels $\boldsymbol{E}_{\text{time}}$. All errors are quantified in terms of the Relative Mean Absolute Error (RMAE).
\begin{equation}\label{error}
\boldsymbol{E}_{\text{sol}}=\frac{\sum_{i=1}^{M}|\boldsymbol{x}_{\boldsymbol{\theta}}(t_i^{\star})-\boldsymbol{x}(t_i^{\star})|}{\sum_{i=1}^{M}|\boldsymbol{x}(t_i^{\star})|}, \boldsymbol{E}_{\text{para}}=\frac{\|\tmmathbf{\theta}-\tmmathbf{\theta}^{\star}\|_1}{\|\tmmathbf{\theta}^{\star}\|_1},  \boldsymbol{E}_{\text{time}}=\frac{\sum_{i=1}^n|\hat{t}_i-t_i|}{\sum_{i=1}^n|t_i|}.
\end{equation}
Here $\tmmathbf{\theta}$ is the estimated parameters of the dynamical system and $\tmmathbf{\theta}^{\star}$ is the ground truth.
The surrogate model in our experiment is a 5-layer MLP with 500 neurons in each layer and SiLU activation function. And we choose AdamW as the optimizer for both phases. All the experiment were done on the Nvidia RTX3090 and Intel Xeon Gold 6130 with 40G RAM. Detailed settings and implementations for the training and evaluation process is provided in Appendix SM3.

\subsection{Simple illustrative systems}
We here apply the reconstruction algorithm to some simple systems with the following form

\[  \left\{ \begin{array}{c}
    \frac{d\tmmathbf{x}}{\tmop{dt}} =\tmmathbf{A}\boldsymbol{f}(\tmmathbf{x}), t \in [0, T]\\
    \tmmathbf{x} (0) = (\tmmathbf{x}_1, \ldots, \tmmathbf{x}_d)^{\top} .
  \end{array} \right. \]

\begin{enumerate}
    \item \textbf{Linear2D equations}:The linear matrix is $A = \begin{bmatrix}
        -0.1&2\\-2&-0.1
    \end{bmatrix}$, the transformation is $\boldsymbol{f}= \mathbb{I}_d$ ,the initial condition is $\tmmathbf{x}(0)=[2,0]$, the time length is $T=10$.
    \item \textbf{Cubic2D equations}:The linear matrix is $A = \begin{bmatrix}
        -0.1&2\\-2&-0.1
    \end{bmatrix}$, the transformation is $\boldsymbol{f}= \begin{bmatrix}
        x_1^3\\
        x_2^3
    \end{bmatrix}$ ,the initial condition is $\tmmathbf{x}(0)=[2,0]$, the time length is $T=10$.
    \item \textbf{Linear3D equations}:The linear matrix is $A = \begin{bmatrix}
        -0.1&2&0\\-2&-0.1&0\\0&0&-0.3
    \end{bmatrix}$, the transformation is $\boldsymbol{f}= \mathbb{I}_d$ ,the initial condition is $\tmmathbf{x}(0)=[2,0,-1]$, the time length is $T=10$.
\end{enumerate}

\begin{figure}[!htbp] 
	\subfigure[]{	\includegraphics[width=0.32\linewidth]{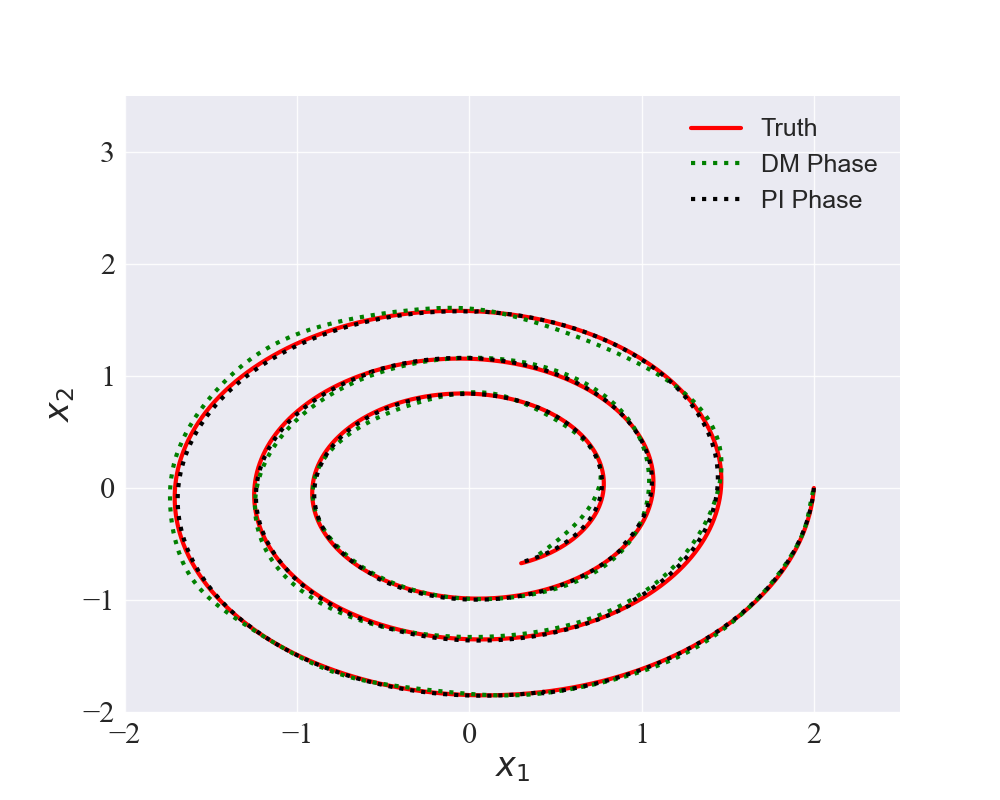}}
	\subfigure[]{\includegraphics[width=0.32\linewidth]{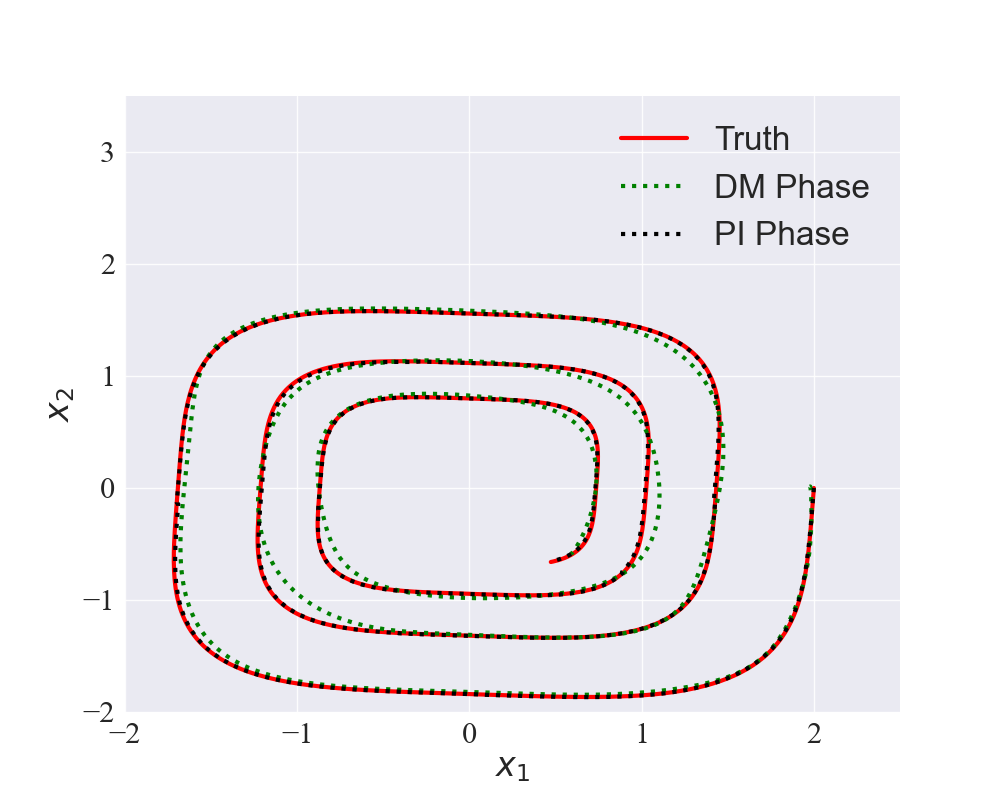}}
        \subfigure[]{		\includegraphics[width=0.32\linewidth]{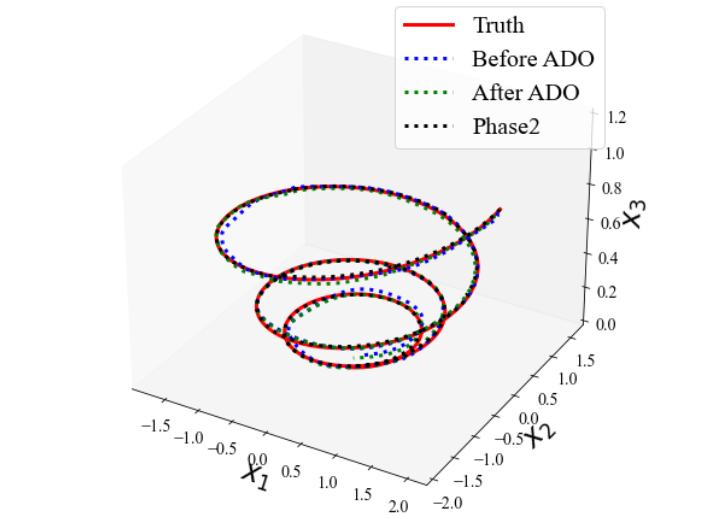}}
        \subfigure[]{		\includegraphics[width=0.32\linewidth]{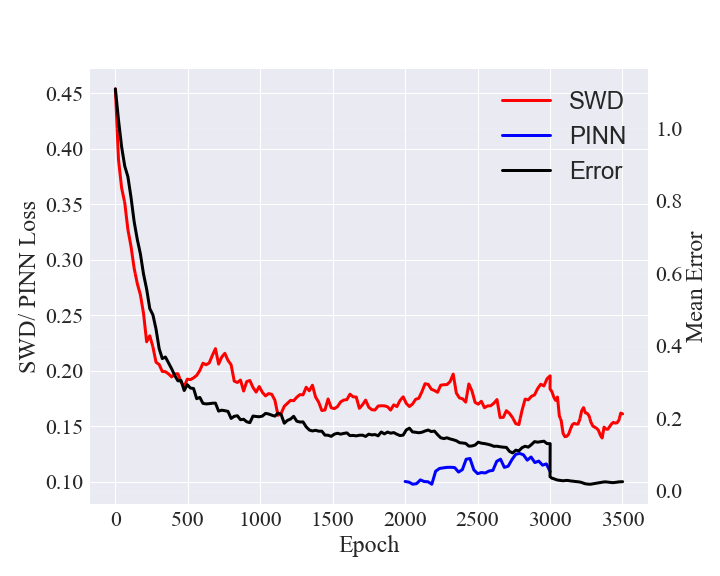}}
        \subfigure[]{		\includegraphics[width=0.32\linewidth]{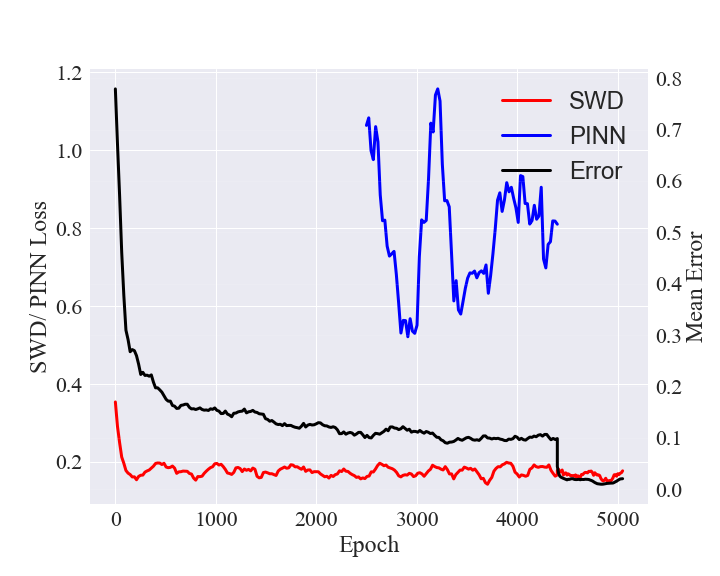}}
        \subfigure[]{		\includegraphics[width=0.32\linewidth]{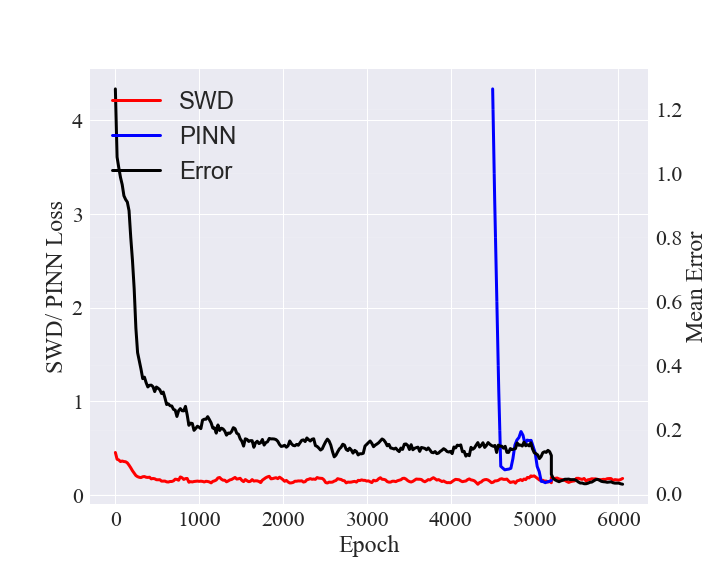}}
        \subfigure[]{		\includegraphics[width=0.32\linewidth]{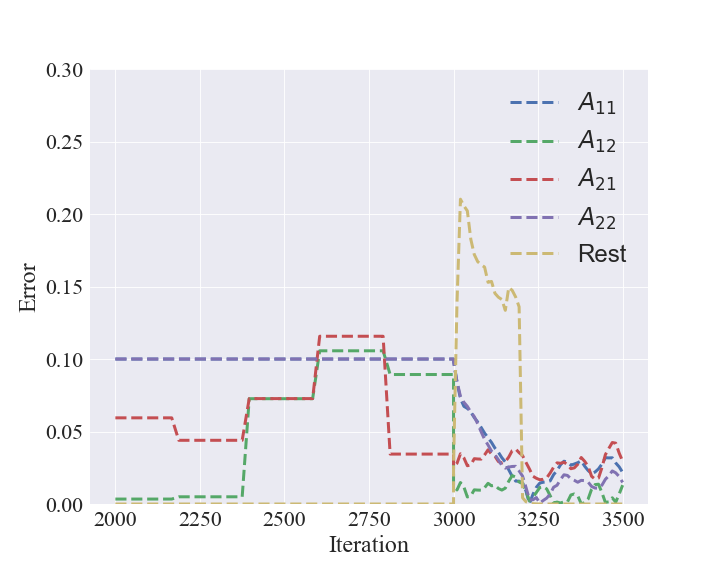}}
        \subfigure[]{		\includegraphics[width=0.32\linewidth]{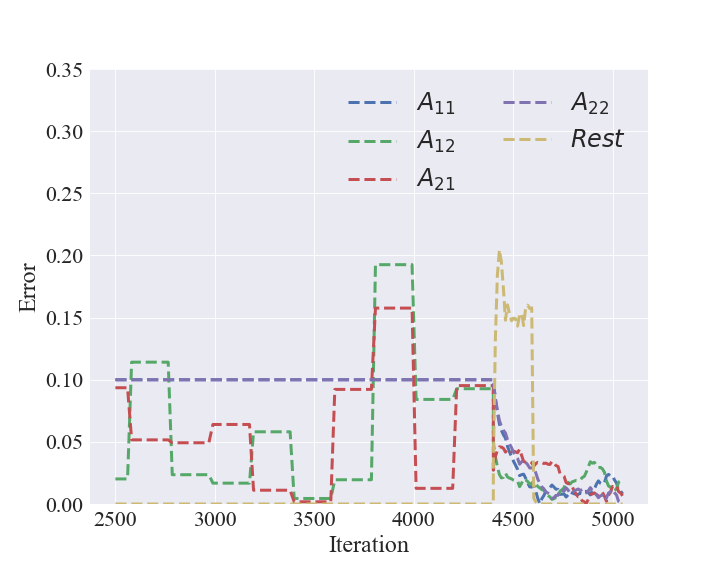}}
        \subfigure[]{		\includegraphics[width=0.32\linewidth]{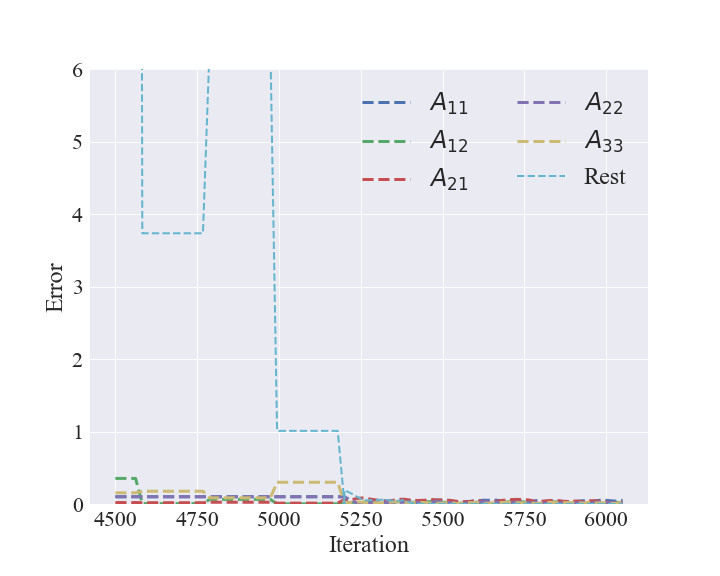}}
        \subfigure[]{		\includegraphics[width=0.32\linewidth]{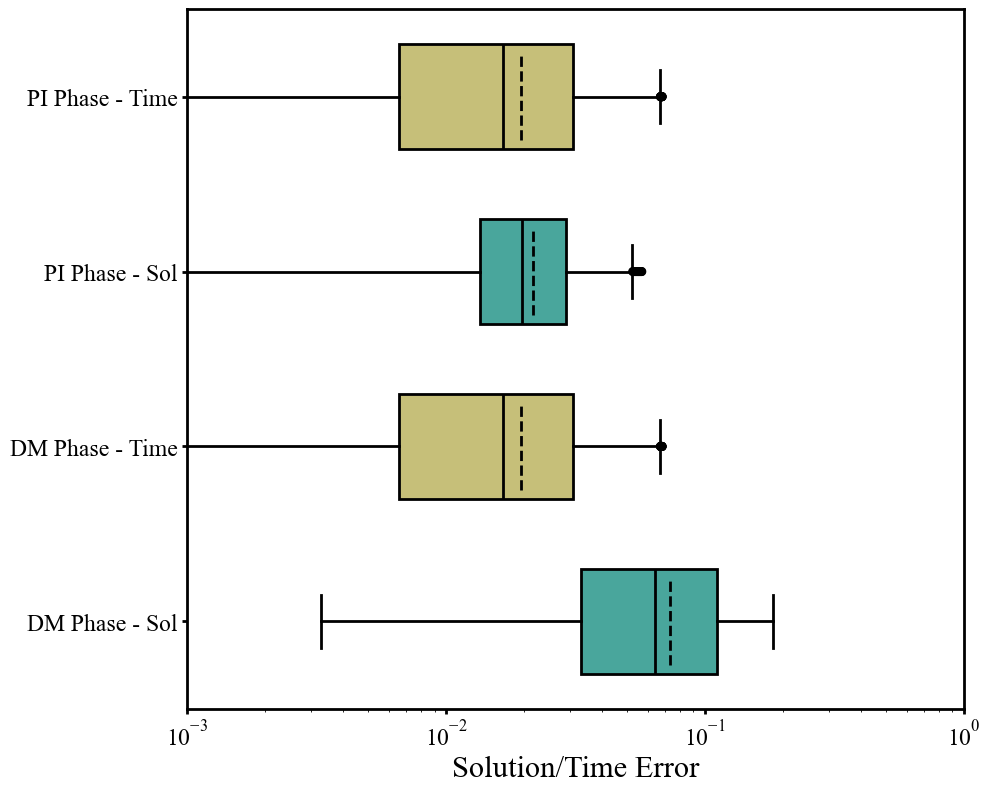}}
        \subfigure[]{		\includegraphics[width=0.32\linewidth]{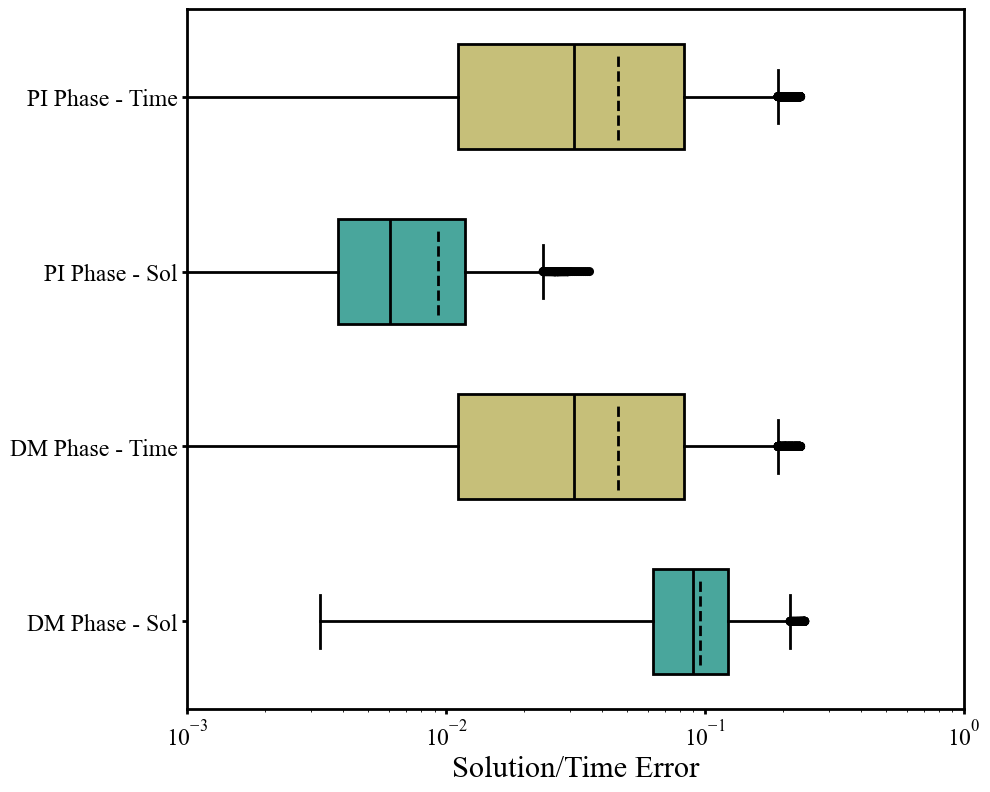}}
        \subfigure[]{		\includegraphics[width=0.32\linewidth]{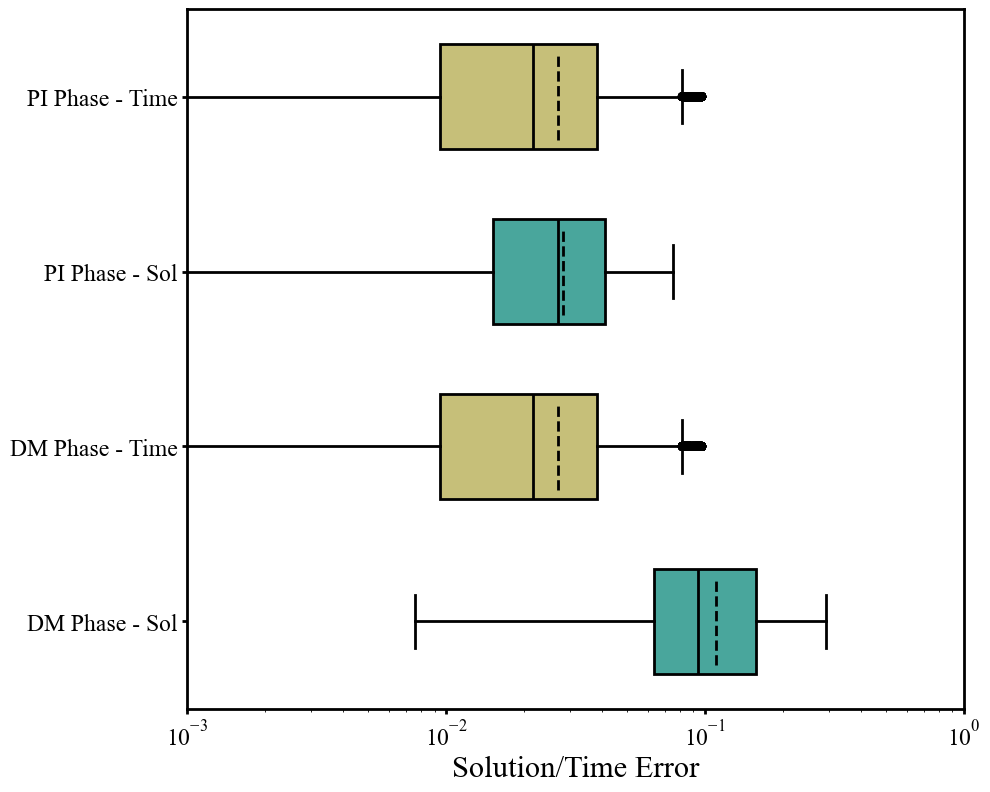}}
	\caption{\textbf{Reconstruction result of illustrative system:} (a)-(c)Comparison between learned curve at different stage and ground truth for all system;(d)-(f) SWD Loss, PINN Loss and Average Error for all system;(g)-(i) The error curve of parameters and the sum of unexisting terms for all system;(j)-(l)The absolute error of learned solution evaluated at uniform grid and the absolute error of reconstructed time label of the dataset.}
	\vspace{0cm}
 \label{fig.illu}
\end{figure}

In these cases, we employ third-order complete polynomials and exponential functions as the basis for the library in both phases. As shown in Fig~\ref{fig.illu}, our method not only reconstructs the solution and temporal labels with exceptional precision but also accurately identifies the underlying system. During the distribution matching phase, at the warm-up stage without the physics-informed regularization term (corresponding to the “Before ADO” stage in Fig~\ref{fig.illu}(a), (b), (c)), the surrogate model is capable of capturing the dynamics' behavior, albeit with minimal accuracy. Following ADO training, the surrogate model accurately approximates the solution and provides an admissible estimate of the system. Although Fig~\ref{fig.illu}(e) and (h) reveal that for higher-order systems such as the Cubic2D problem, the system estimation during the distribution matching phase is unstable and lacks sufficient precision, using its results as an initial guess enables the parameter identification phase to rapidly converge to highly accurate estimates of both the form and parameters of the underlying dynamics, with the solution being fitted with optimal accuracy. This finding underscores the necessity of the two-stage learning process.

\begin{table}[htbp]
\tabcolsep=0.1cm
    \centering
    \begin{tabular}{c|c|c|c|c|c}
        \toprule
        Settings &$||\boldsymbol{\mathbb{X}}||_{rms}$& $\boldsymbol{E}_{\text{sol}}\downarrow$ & $\boldsymbol{E}_{\text{para}}\downarrow$ & $\boldsymbol{E}_{\text{time}}\downarrow$ & $A$ \\
        \hline
        Linear2D & 1.26 &\textbf{1.4}$\%$ (8.4$\%$)& \textbf{1.9}$\%$ (7.7$\%$)&  \textbf{0.056}$\%$ (0.39$\%$)& $\begin{bmatrix}
           -0.12& 1.97\\-2.01&-0.08
        \end{bmatrix}$ \\
        &&&&&\\
        Cubic2D &1.13 &\textbf{0.6}$\%$ (7.0$\%$)& \textbf{0.8}$\%$ (9.2$\%$)& \textbf{0.4}$\%$ (0.9$\%$)& $\begin{bmatrix}
           -0.11& 1.99\\-2.02&-0.99
        \end{bmatrix}$ \\
        &&&&&\\
        Linear3D & 1.31 & \textbf{1.5}$\%$ (5.7$\%$)& \textbf{3.0}$\%$ (6.8$\%$)&  \textbf{0.13}$\%$ (0.53$\%$)& $\begin{bmatrix}
           -0.15&  1.97& 0\\-2.01&-0.07&0\\
           0&0&-0.28
        \end{bmatrix}$  \\
        \bottomrule
    \end{tabular}
    \caption{\textbf{The results of illustrative examples:} The symbol \(\|\cdot\|_{rms}\) denotes the root mean square of the observational data, reflecting the data scale. We recorded the values of all three metrics \(\boldsymbol{E}_{\text{sol}}\), \(\boldsymbol{E}_{\text{para}}\), and \(\boldsymbol{E}_{\text{time}}\) in both phases. The metrics in brackets are computed using the learned neural solution from the distribution matching phase and the final results are marked in bold. All the inactive terms in library are successfully removed during training in four experiment, so the Parameters row describes the identified systems. } 
    \label{tab.1d5}
\end{table}

\subsection{Benchmark Examples}
In the instances delineated below, we execute our algorithm on a selection of challenging ODE systems derived from real-world models.
\begin{enumerate}
    \item Lorenz equations represent a early mathematical model for characterizing atmospheric turbulence.This system exhibits chaotic behavior over the long time horizon, resulting in intricate intersecting trajectories and convoluted spiraling within the phase space
    \item The Lotka-Volterra (LV) equations depict the dynamics of prey and predator populations in natural ecosystems and have found applications in real-world systems such as disease control and pollution management.  In this study, we couple two independent LV equations(LV4D) to evaluate the efficacy of our algorithm in addressing high-dimensional problems.
    \item The Duffing oscillator serves as a model for specific damped and driven oscillators, demonstrating long-term chaotic behavior.
\end{enumerate}
\begin{table}[htbp]
\centering
\begin{tabular}{c|c|c}
\hline
 Name& ODE& Parameters \\ \hline
 Lorenz& $
    \left\{
    \begin{array}{l}
    \frac{dx_1}{dt} = \sigma (x_2-x_1)\\
    \frac{dx_2}{dt} = x_1 (\rho-x_3)-x_2\\
    \frac{dx_3}{dt} = x_1 x_2-\beta x_3
  \end{array}
    \right.
$ & \makecell[c]{$T=3$\\$[\sigma,\rho,\beta] = [10,28,\frac{8}{3}]$\\$\boldsymbol{x_0}=(10,-10,20)^\top$}\\ \hline
 Lotka-Volterra4D&$
\left\{
\begin{array}{l}
\frac{dx_1}{dt} = \alpha_1 x_1-\beta_1 x_1 x_2\\
\frac{dx_2}{dt} = \beta_1 x_1 x_2-2\alpha_1 x_2\\
\frac{dx_3}{dt} = \alpha_2 x_3-\beta_2 x_3 x_4\\
\frac{dx_4}{dt} = \beta_2 x_3 x_4-2\alpha_2 x_4
\end{array}
\right.
$ & \makecell[c]{$T=11$\\$[\alpha_1,\alpha_2,\beta_1,\beta_2] = [1,1,3,5]$\\$\boldsymbol{x_0}=(1,2,2,0.5)^\top$} \\ \hline
 Duffing& $
\left\{
\begin{array}{l}
\frac{dx_1}{dt} = \alpha x_2\\
\frac{dx_2}{dt} = -\gamma x_1 - \rho x_2 -\beta x_1^3
\end{array}
\right.
$ & \makecell[c]{$T=11$\\$[\alpha,\gamma,\rho,\beta] = [1,0.1,0.2,1]$\\$\boldsymbol{x_0}=(0,2)^\top$} \\
\hline
\end{tabular}
\caption{\textbf{Specifications of ODE examples.}}
\end{table}

\begin{figure}[htbp] 
	\subfigure[]{
\includegraphics[width=0.3\textwidth]{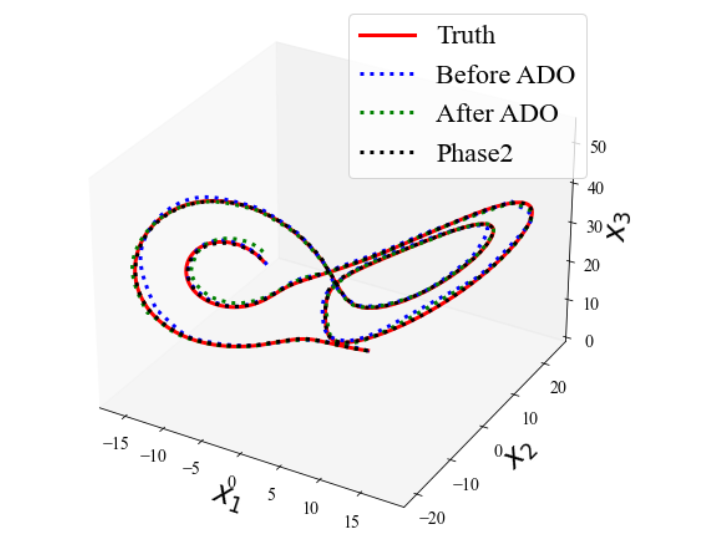}
}
       	\subfigure[]{ 
\includegraphics[width=0.3\textwidth]{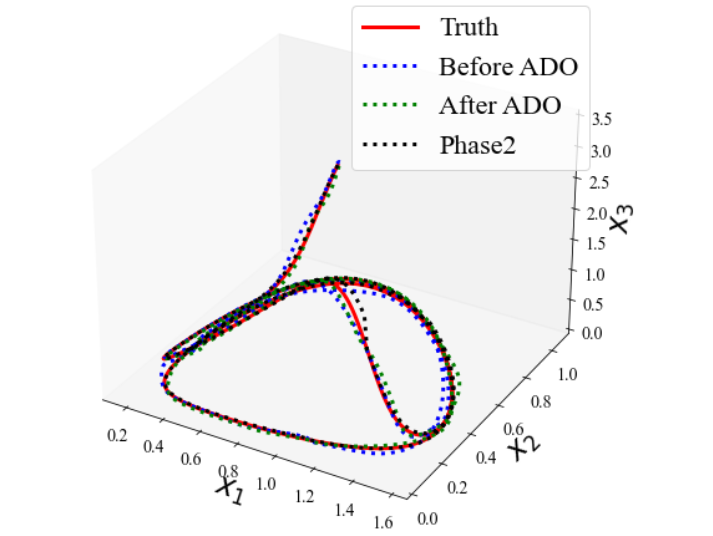}
}
\subfigure[]{ 
\includegraphics[width=0.3\textwidth]{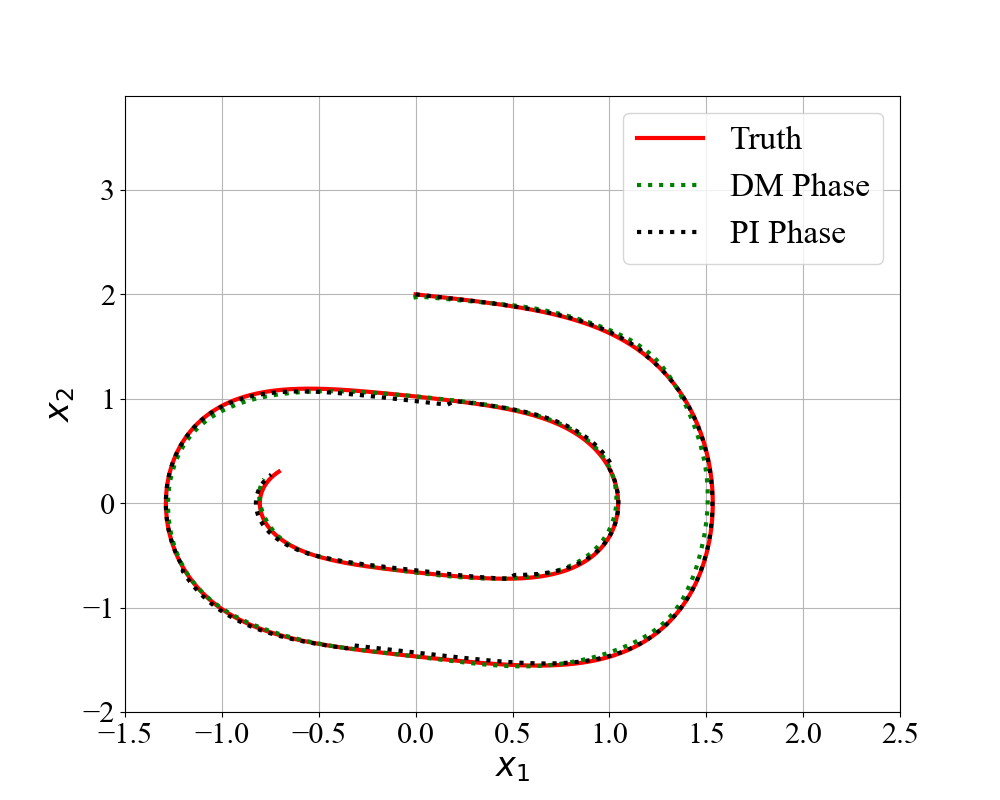}
}

\subfigure[]{		\includegraphics[width=0.3\textwidth]{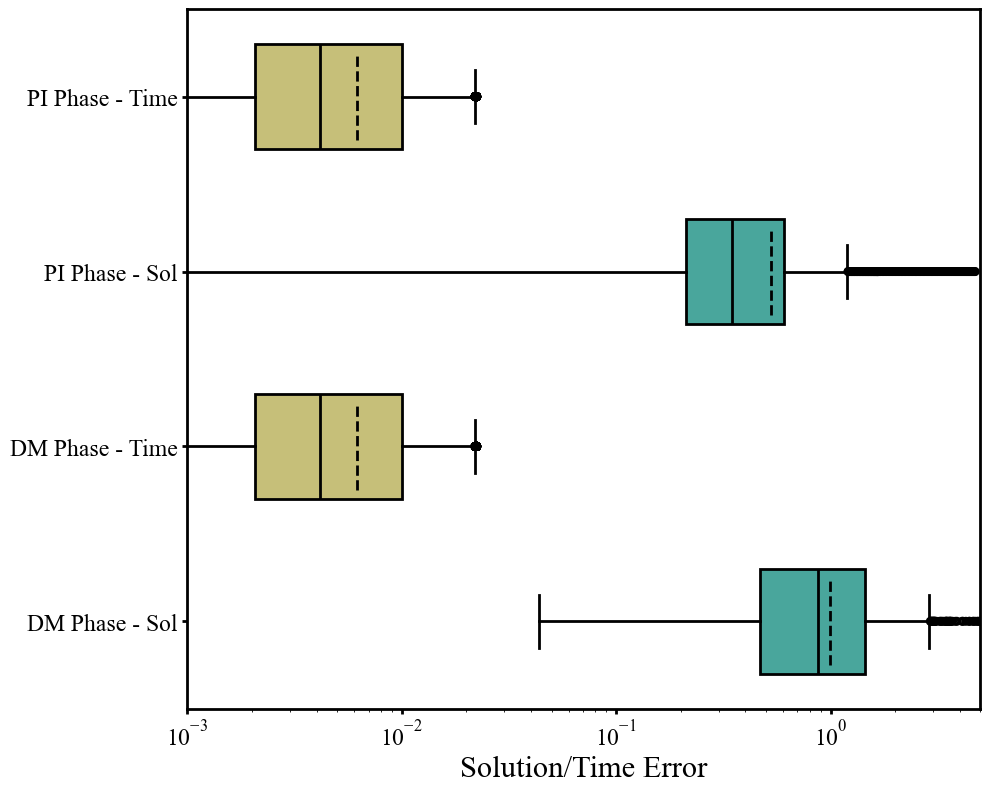}}
        \subfigure[]{		\includegraphics[width=0.3\textwidth]{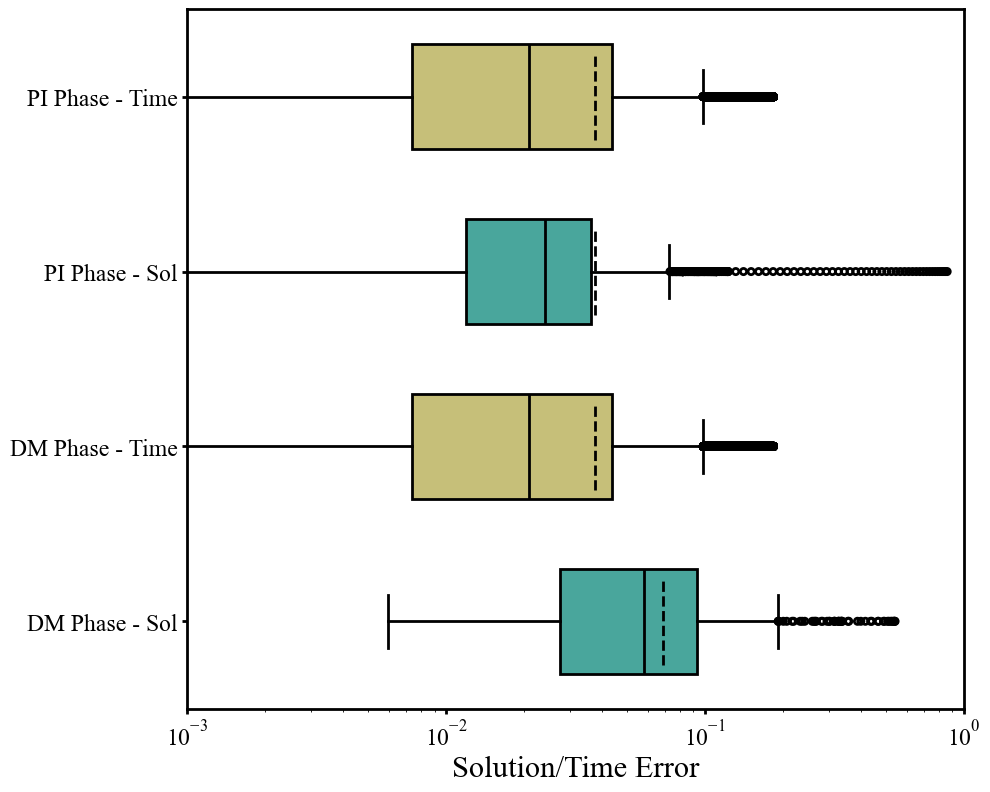}}
        \subfigure[]{		\includegraphics[width=0.3\textwidth]{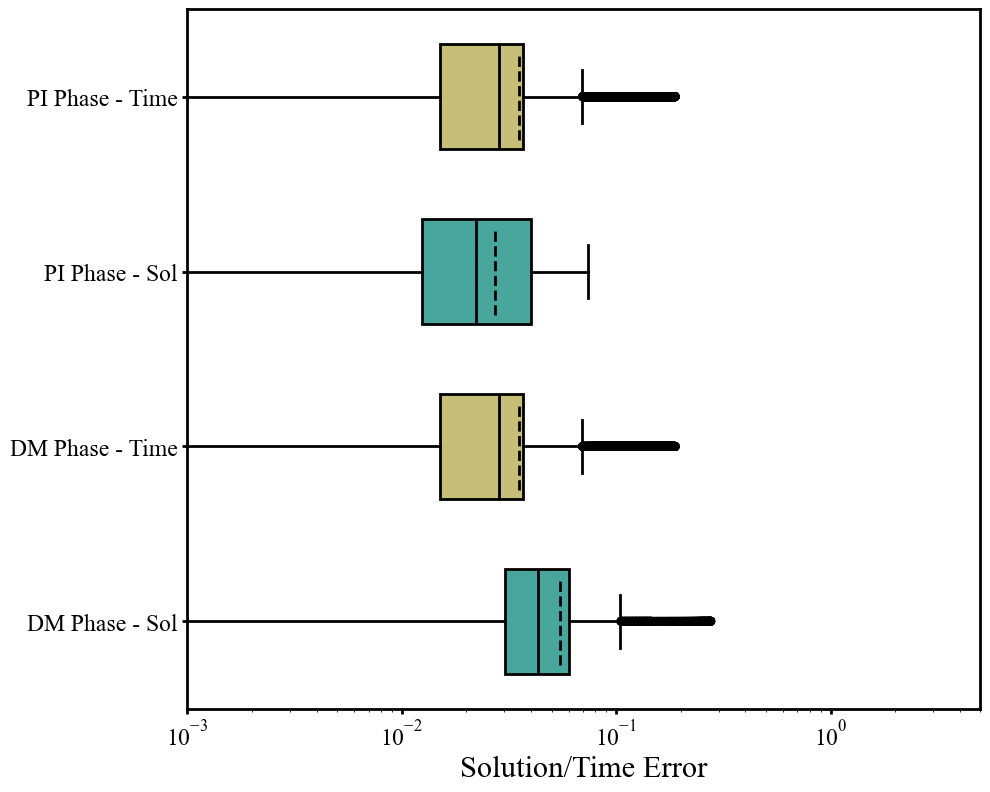}}
	\caption{\textbf{Reconstruction result of benchmark system:}Reconstructing trajectories of four systems at different training stage:(a)Lorenz system; (b)LV4D system, plot the curve of first 3 dimension;(c)Duffing system. The point-wise error of learned solution function and time label of three training stage: (a)Lorenz system; (b)LV4D system; (c)Duffing system.}
	\vspace{0cm}
	\label{Exp1.1}   
\end{figure}

As shown in Table \ref{tab:ex1} and Fig \ref{Exp1.1}, our approach consistently achieves high accuracy across all three benchmarks—encompassing both chaotic and high-dimensional systems—thus demonstrating the wide applicability of our method to diverse ODEs. During the distribution matching phase, the surrogate model approximates the true dynamic system with an RMAE of less than $10\%$ for both the solution function and the system parameters. Notably, at this stage, the reconstructed time labels are highly accurate, with an RMAE below $1\%$. Furthermore, in the parameter identification phase, after only a few iterations, the solution accuracy across all systems improves by a factor of 4 to 10, and all active terms are successfully identified without redundancy. Ultimately, the reconstructed time labels are highly accurate, with an RMAE below $0.5\%$. It should be noted, however, that despite the model's accurate identification, the reconstruction error for both the solution and time labels may remain high over certain intervals when two trajectory segments are close to each other. This phenomenon is evident in Fig~\ref{Exp1.1}(a)–(c) for the Lorenz and LV4D systems, whose trajectories nearly coincide over distinct intervals.

To illustrate the precision of our approach, we compare it with the Principal Curve method—a conventional manifold parametrization technique. In our evaluation, the index produced by the Principal Curve algorithm \cite{hastie1989principal} serves as the reconstructed time label. Specifically, we apply the Principal Curve algorithm to each segmented trajectory piece, designating the estimated initial condition as the starting point of each curve. The resulting indices for each data point are then normalized to the interval \([T_l, T_{l+1}]\) for each segment \(\mathbb{X}_l\). Our results demonstrate that, for the same dataset, our method achieves over a 30-fold improvement in accuracy compared to the traditional Principal Curve approach, which notably fails to resolve time labels between intervals. This limitation is partly attributable to the underlying parametrization principle of the Principal Curve algorithm, which assigns indices based solely on the Euclidean distance between points—a measure that often does not capture the true velocity of the curve as determined by the observation distribution.

\begin{table}[htbp]
\tabcolsep=0.1cm
\centering
\begin{tabular}{c|c|c|c|c|c|c}
    \toprule
    Settings & $||\boldsymbol{\mathbb{X}}||_{rms}$ & $\boldsymbol{E}_{\text{sol}}\downarrow$ & $\boldsymbol{E}_{\text{para}}\downarrow$ & \multicolumn{2}{c|}{$\boldsymbol{E}_{\text{time}}\downarrow$} & Parameters \\
    \cline{5-6}
             &                                   &                                  &                                   & Our                   & PC                  &            \\
    \hline
    Lorenz  & 25.36  & \textbf{1.5}\% (2.8\%)     & \textbf{0.58}\% (0.65\%)     & \textbf{0.10}\% (0.40\%)  & 25.9\%     & $\begin{bmatrix}
        \sigma \\ 
        \rho\\
        \beta
    \end{bmatrix} =\begin{bmatrix}
        9.94 \\ 
        27.89\\
        2.59
    \end{bmatrix}$ \\
   &&&&&& \\
    LV4D   & 1.12   & \textbf{2.3}\% (4.2\%)      & \textbf{4.3}\% (5.6\%)      & \textbf{0.31}\% (0.69\%)  & 13.0\%     & $\begin{bmatrix}
        \alpha_1\\
        \alpha_2\\
        \beta_1\\
        \beta_2
    \end{bmatrix} =\begin{bmatrix}
        0.97 \\ 
        0.98\\
        2.93\\
        4.65
    \end{bmatrix}$ \\
    &&&&&& \\
    Duffing & 1.17   & \textbf{1.63}\% (3.88\%)   & \textbf{2.03}\% (8.53\%)   & \textbf{0.20}\% (0.30\%)  & 19.4\%    & $\begin{bmatrix}
        \alpha\\
        \gamma\\
        \rho\\
        \beta\\
    \end{bmatrix} =\begin{bmatrix}
        0.99 \\ 
        0\\
        0.21\\
        1.06
    \end{bmatrix}$ \\
    \bottomrule
\end{tabular}
\caption{\textbf{Benchmark Results:} The symbol \(||\cdot||_{rms}\) denotes the root mean square of the observational data. The metrics \(\boldsymbol{E}_{\text{sol}}\) and \(\boldsymbol{E}_{\text{para}}\) are computed in the first phase (using the learned neural solution), while the two columns under \(\boldsymbol{E}_{\text{time}}\) denote the results of our method (Our) and the traditional Principal Curve (PC) method, respectively. \textbf{Bold} indicates the best performance.}
\label{tab:ex1}
\end{table}

\begin{figure}[htbp] 
	\subfigure[]{
 \centering
 \begin{minipage}[b]{0.48\textwidth}
\includegraphics[width=1\linewidth]{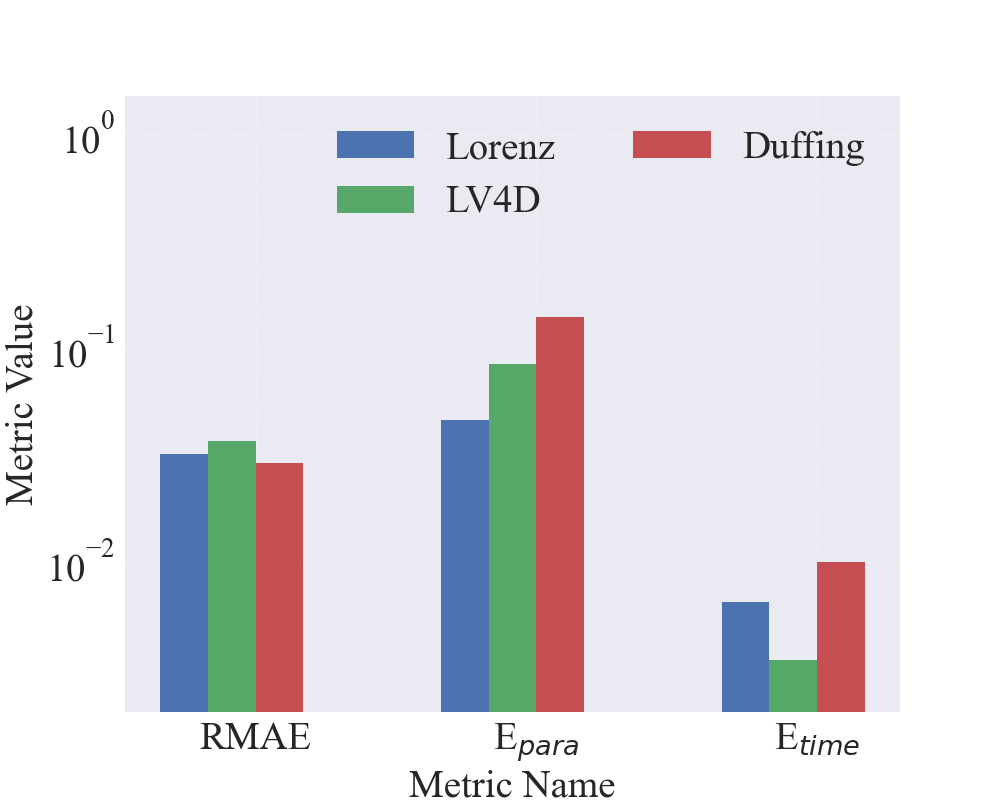}
\end{minipage}}
       	\subfigure[]{ \centering
 \begin{minipage}[b]{0.48\textwidth}
\includegraphics[width=1\linewidth]{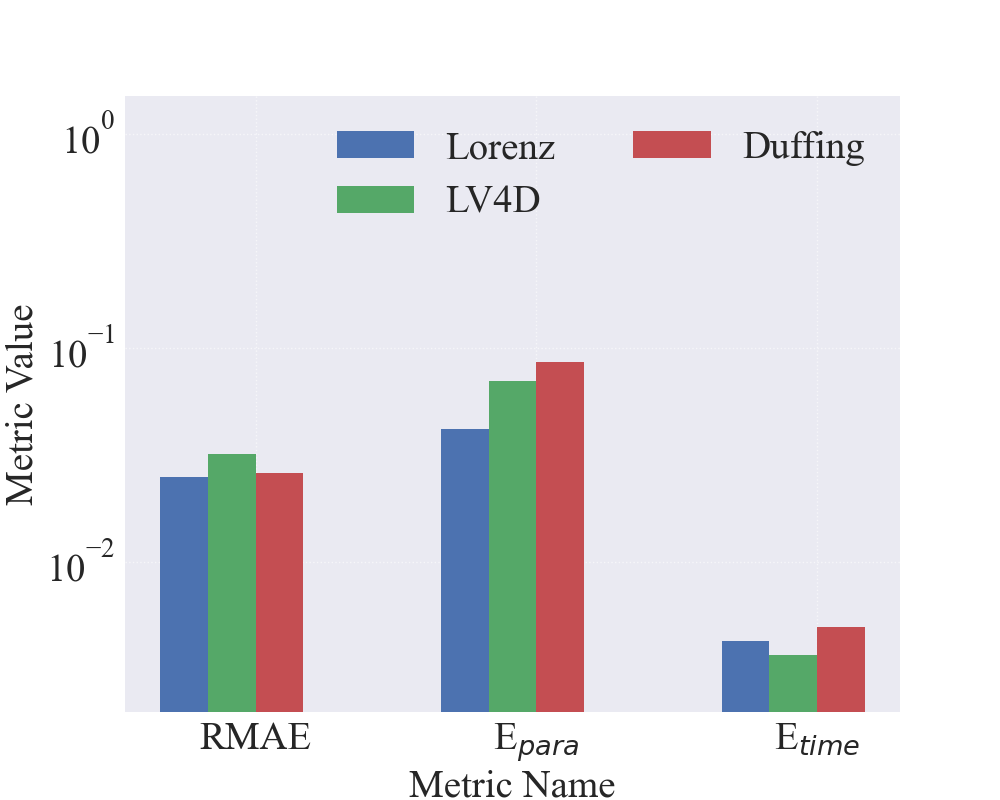}
\end{minipage}}
	\caption{Evaluation of reconstruction results from data with truncated normal distribution of observation time.(a):The evaluation metrics of reconstruction results in distribution matching phase;(b):The evaluation metrics of reconstruction results in parameter identification phase}
	\vspace{0cm}
	\label{Exp3.1}   
\end{figure}

We further investigate the robustness of our method for data with different observation distribution. Here we truncate the normal distribution $\mathcal{N}\left(\frac{T}{2},\left(\frac{T}{3}\right)^2\right)$ in $[0,T]$ and treat it as the observation distribution. The results of four benchmark systems are shown in Fig \ref{Exp3.1} . Overall, our method presents a stable performance w.r.t different observation distribution for reconstructing the solution and time label. 
\subsection{Noisy data}
In this part, we consider the performance of our method on the noisy data. To this end, a noise ratio $\sigma_{NR}$ is specified, and the observation data is blurred by 
\begin{equation}
    \hat{\boldsymbol{\mathbb{X}}}= \boldsymbol{\mathbb{X}}+\epsilon
\end{equation}
where $\boldsymbol{\mathbb{X}}$ is the simulated noise-free observation and $\epsilon$ is the white noise with variance $\sigma^2$, where $
    \sigma = \sigma_{NR}\times||\boldsymbol{\mathbb{X}}||_{rms} = \sigma_{NR}\times\frac{1}{n}\sum_{i=1}^n\left(\sum_{j=1}^d \boldsymbol{x_{i,j}}^2\right)^{\frac{1}{2}}$

In addition to the single-trajectory configuration described above, we also present results from the observation of multiple short trajectories. This setting can be regarded as a precise segmentation of a long trajectory under noise-free initial conditions, thereby eliminating the need for explicit segmentation. We assess the robustness on four benchmark systems using noise levels of $1\%$ and $3\%$ in the single-trajectory setting and $3\%$ and $5\%$ in the multiple-trajectory setting. 

\begin{figure}[!htbp] 
	\subfigure[]{
\includegraphics[width=0.3\linewidth]{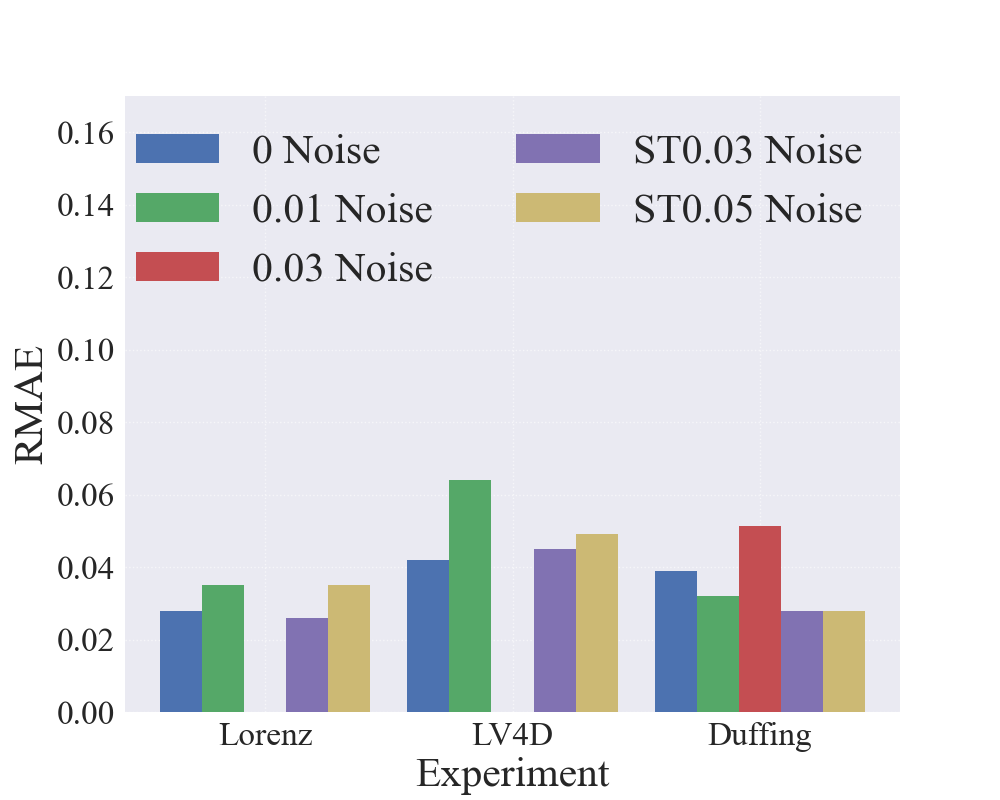}
}
	\subfigure[]{ \centering
\includegraphics[width=0.3\linewidth]{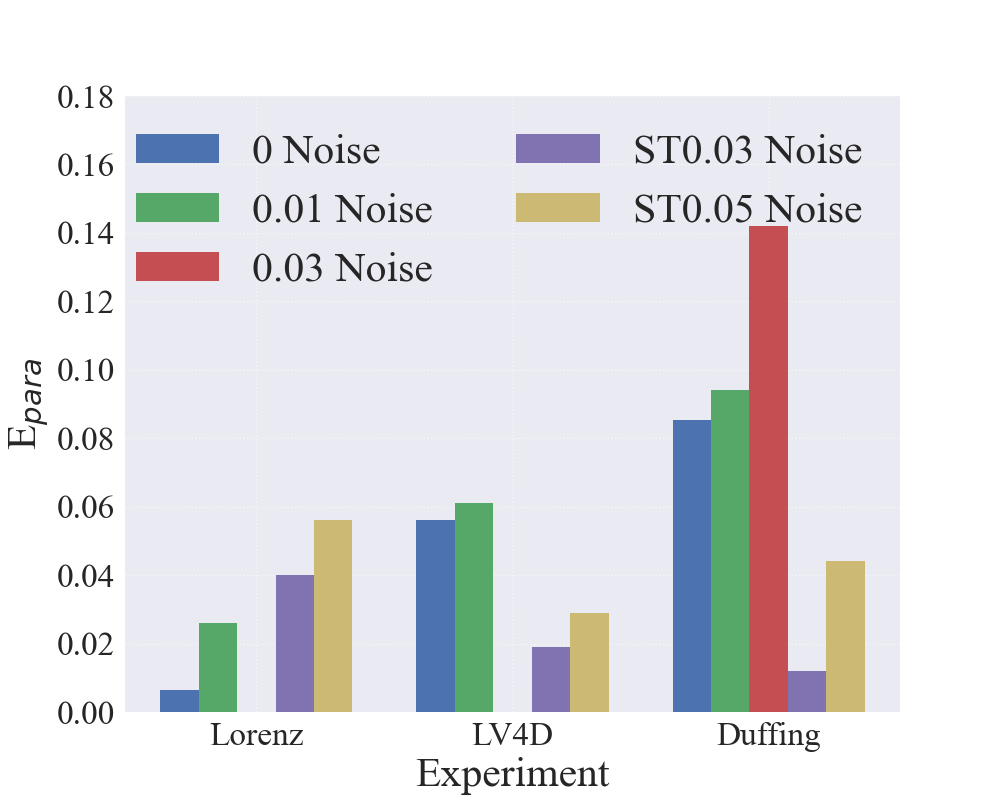}
}
	\subfigure[]{ \centering
\includegraphics[width=0.3\linewidth]{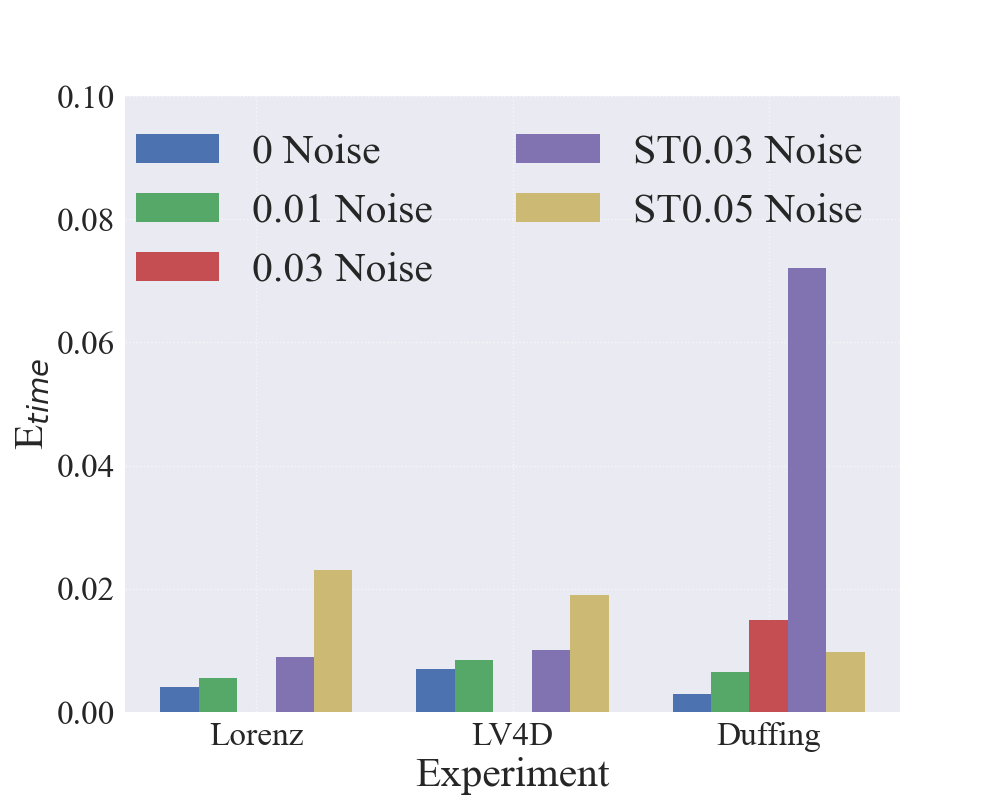}
}

        \subfigure[]{ \centering
\includegraphics[width=0.3\linewidth]{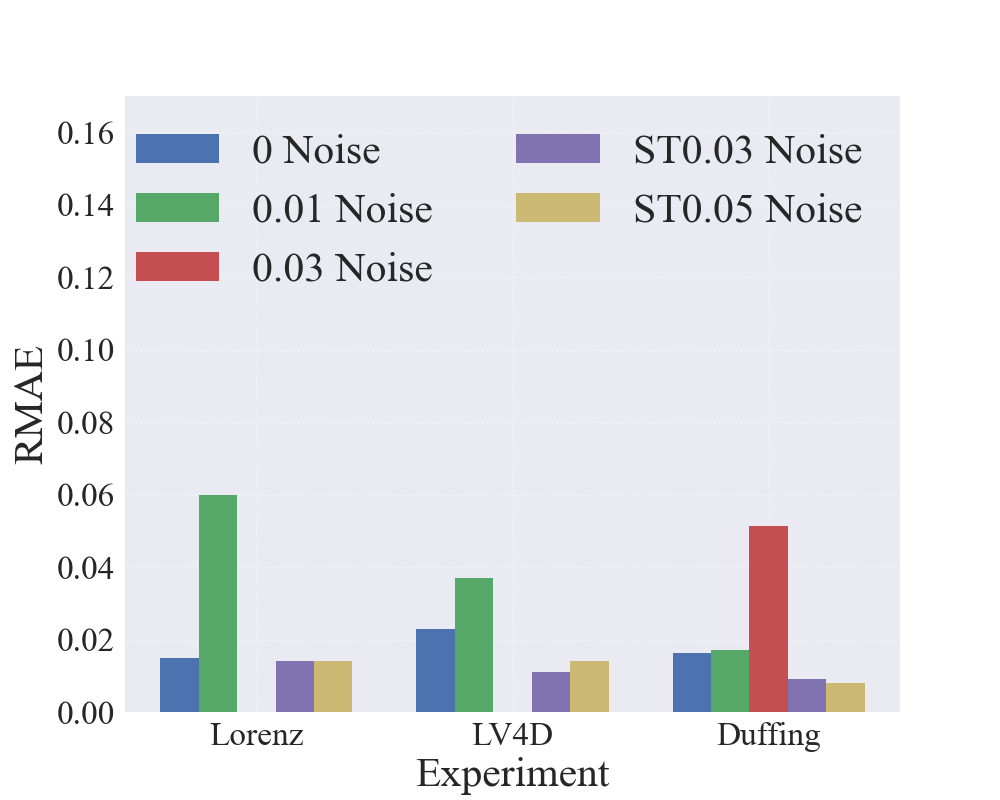}
}
        	\subfigure[]{ \centering
\includegraphics[width=0.3\linewidth]{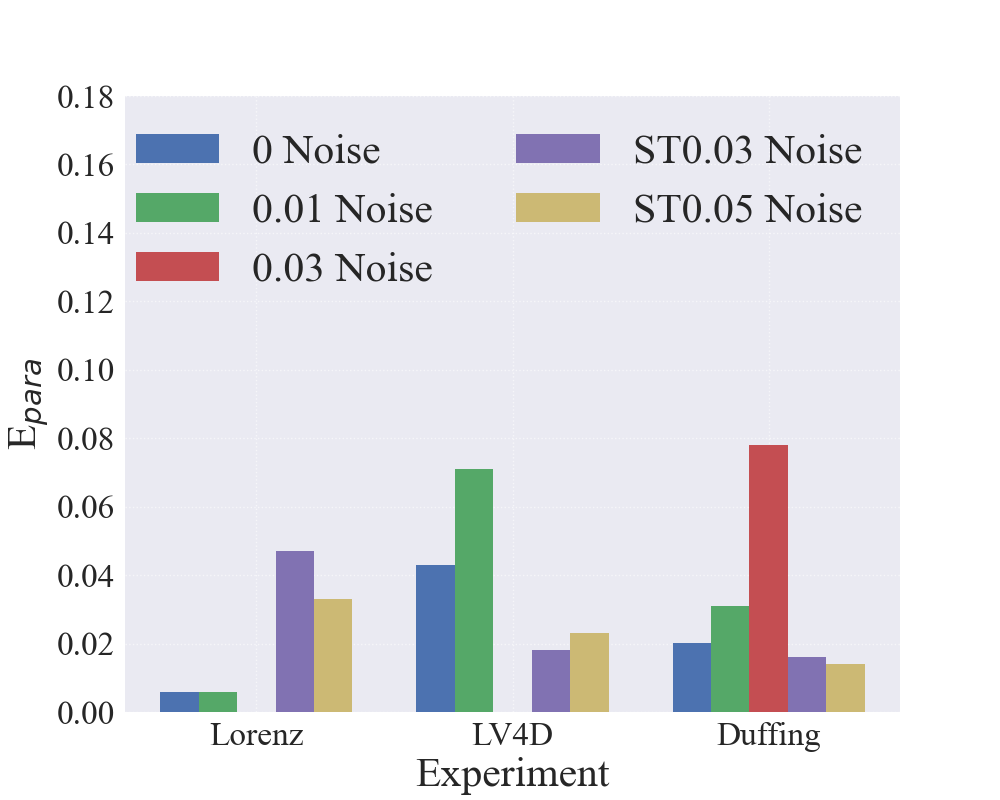}
}
        	\subfigure[]{ \centering
\includegraphics[width=0.3\linewidth]{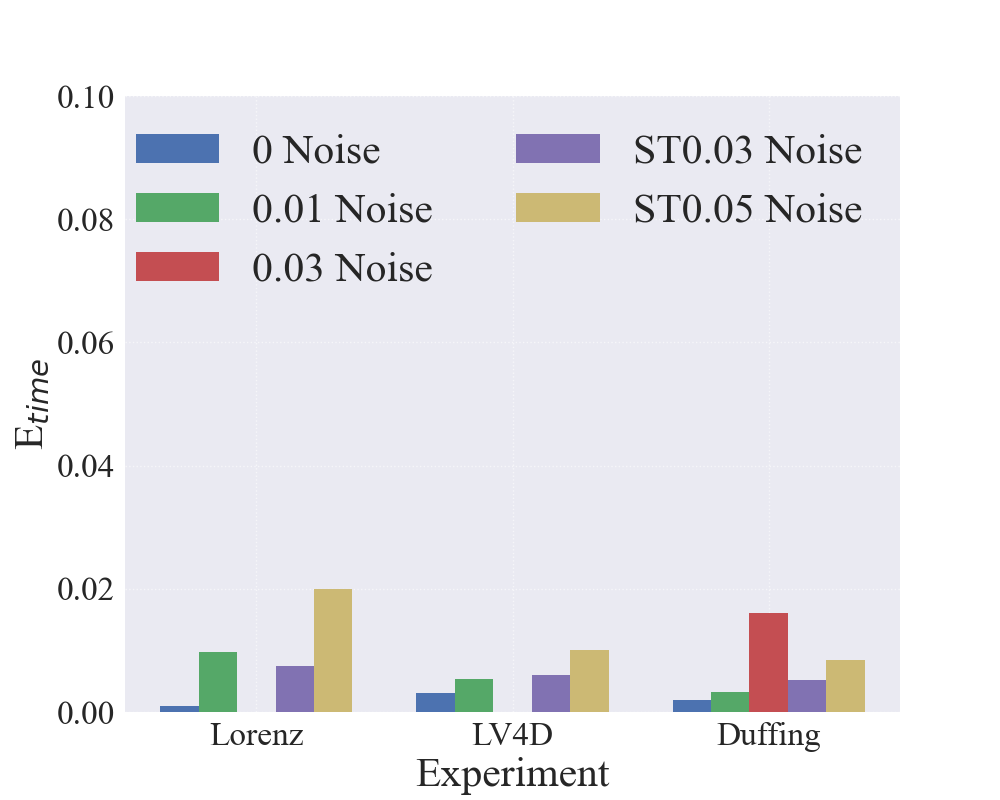}
}
	\caption{\textbf{Evaluation of reconstruction results from noisy data in two trajectory type and four systems:}(a)-(c):The RMAE of learned solution, identified parameters and reconstructed time labels in the distribution matching phase;
 (d)-(f):The RMAE of learned solution, identified parameters and reconstructed time labels in the parameter identification phase. The data of Lorenz and LV4D systems for $3\%$ noise is missing because the segmentation step failed to provide non-overlapping trajectory pieces.}
	\vspace{0cm}
	\label{Exp2.1}   
\end{figure}

As shown in Fig~\ref{Exp2.1}, the two-phase learning process enables us to achieve system parameter identification and time label reconstruction errors below $10\%$ for noisy data, thereby facilitating accurate reconstruction. Although the surrogate model obtained during the distribution matching phase exhibits substantially higher errors under noisy conditions than in the noise-free case, subsequent parameter identification markedly enhances reconstruction accuracy across all settings, including both single-trajectory and multi-trajectory scenarios. Notably, at high noise levels such as $3\%$ for the setting of single trajectory, different part of the trajectory may overlap, leading to segmentation errors; if a cluster comprises two disjoint segments with distinct initial conditions, the single-trajectory assumption is invalidated.

\subsection{General Hamiltonian system out of basis  }
To illustrate the performance of the basis expansion, we investigate the scenario of the
complex force term without complete expansion of the basis.
 Namely, in this subsection, we try to reveal the hidden dynamics from
data of the famous Pendulum system:
\begin{equation}
\left\{
\begin{array}{l}
\frac{dx_1}{dt} = \alpha x_2\\
\frac{dx_2}{dt} = \beta \sin(x_1)\\
\end{array}
\right.
\end{equation}
where the initial condition is $\boldsymbol{x_0}=(1,0.1)^\top$, the time length is $T=8$ and the parameters are $[\alpha,\beta] = [1,1]$. we note that the forcing term $\sin(x_1)$ cannot be accurately represented by a finite polynomial basis. Moreover, in this case the Hamiltonian $H(x_1,x_2) = \frac{1}{2}x_2^2+\cos(x_1)$ is a conserved quantity describing the total energy of a Hamiltonian system.

We evaluate our method using two basis sets: a larger basis,
  that encompasses all the terms present in the pendulum system, and a smaller 3rd-order complete polynomial library, 
  $$\tmmathbf{\phi}_1 (\tmmathbf{x}) = [Poly(3),\sin(x_1),\sin(x_2),\cos(x_1),\cos(x_2)],\tmmathbf{\phi}_2 = [Poly(3)]$$
  Here Observations at five distinct noise levels are considered to provide a more comprehensive evaluation.
\begin{figure}[!htbp] 
	\subfigure[]{	\includegraphics[width=0.32\linewidth]{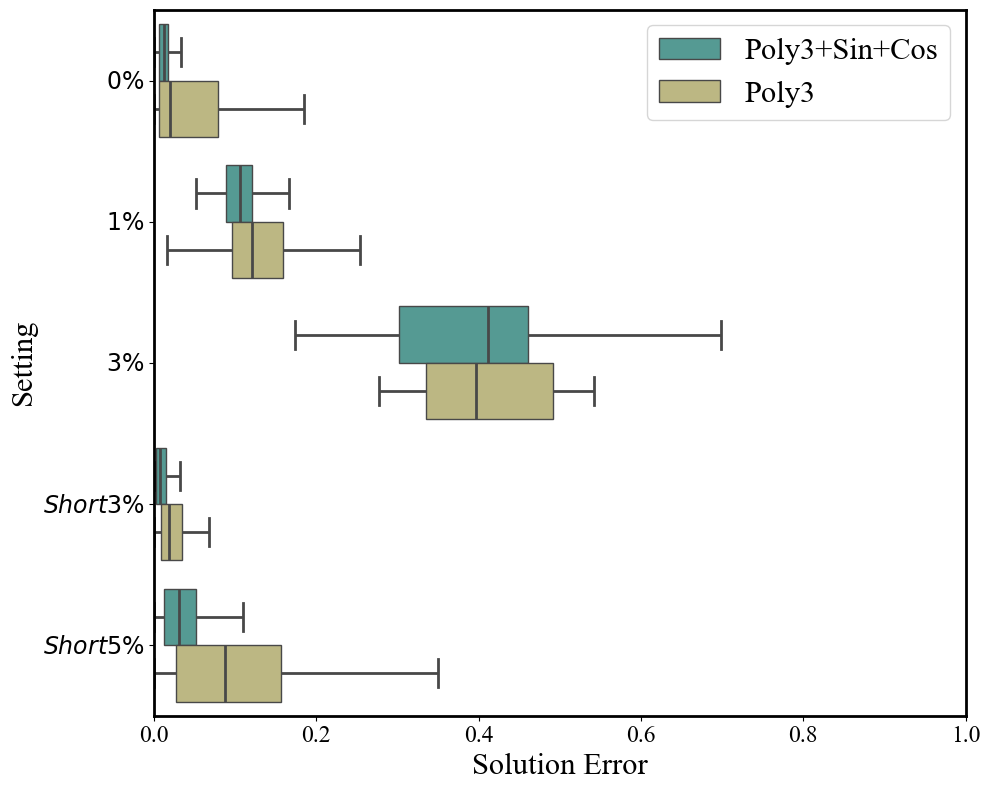}}
	\subfigure[]{\includegraphics[width=0.32\linewidth]{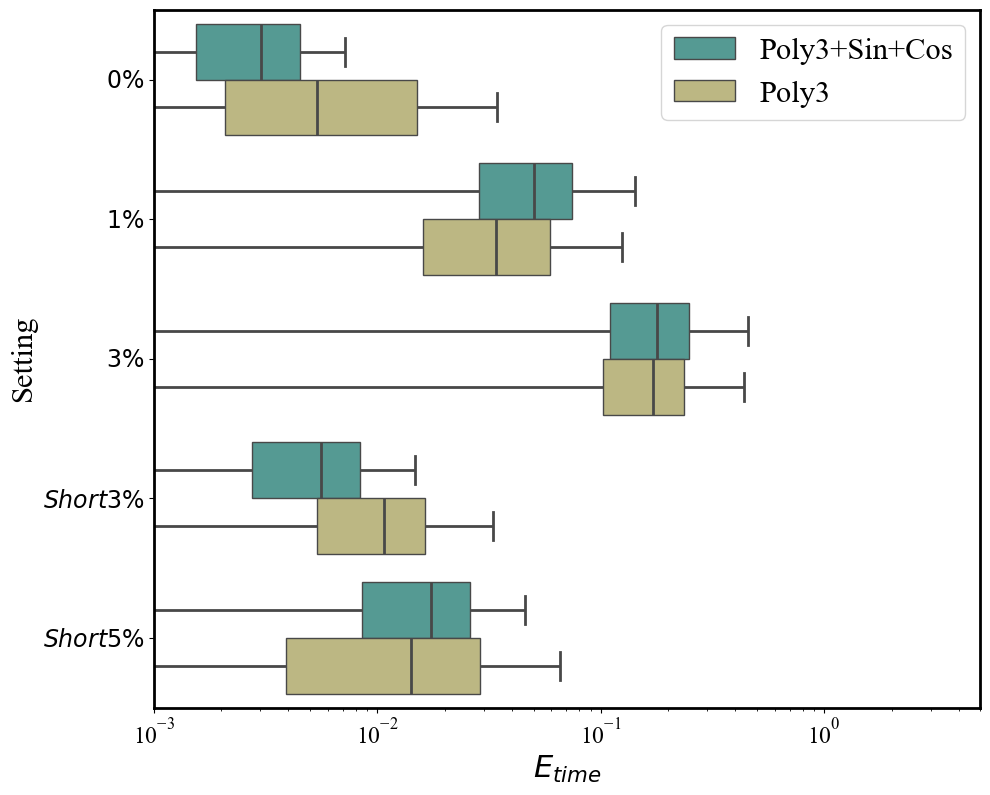}}
        \subfigure[]{		\includegraphics[width=0.32\linewidth]{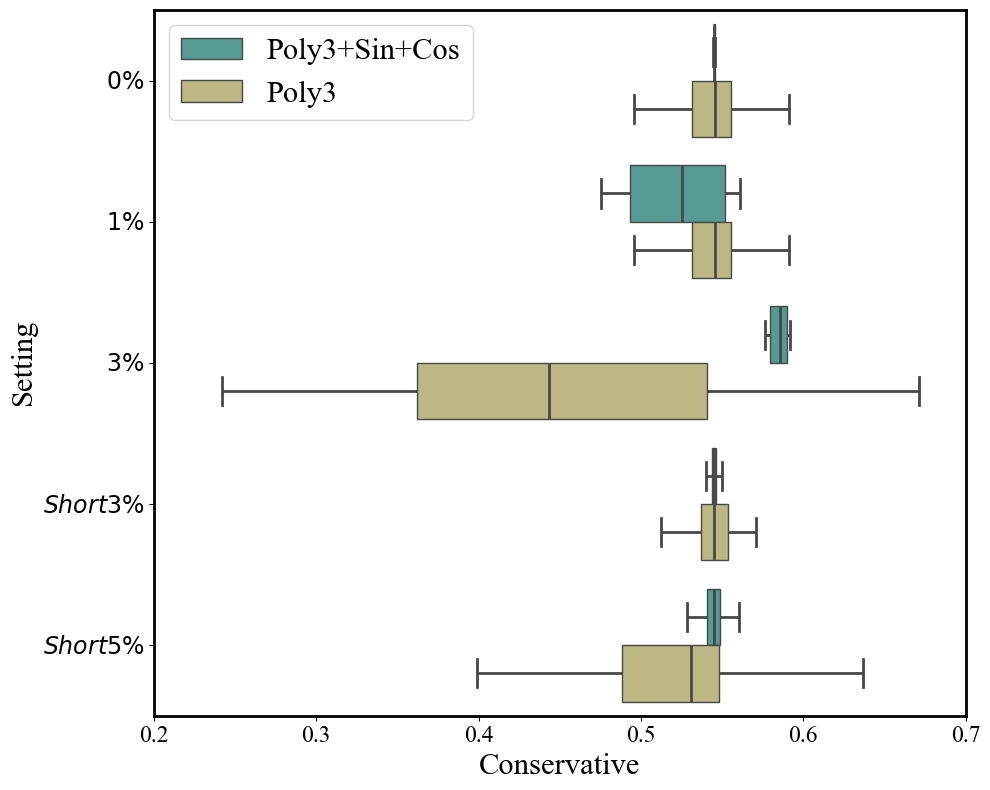}}
	\caption{\textbf{Reconstruction result of Pendulum system:}(a)Violin plot of absolute error of learned solution function for two basis setting in 5 different noise level;(b)Violin plot of absolute error of reconstructed time label for two basis setting in 5 different noise level;(c)Voilin plot of Hamiltonian of learned dynamical system in the parameter identification phase}
	\vspace{0cm}
 \label{fig.sin_and_not}
\end{figure}

\begin{figure}[htbp] 
	\subfigure[]{
 \centering
 \begin{minipage}[b]{0.48\textwidth}
\includegraphics[width=1\linewidth]{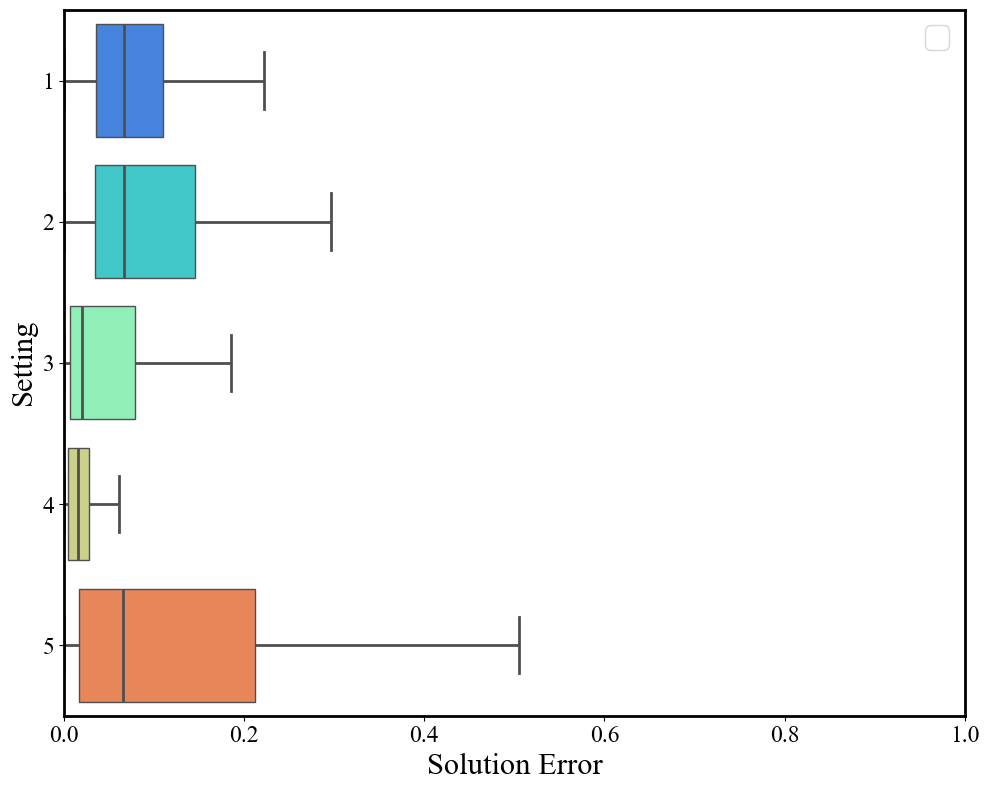}
\end{minipage}}
       	\subfigure[]{ \centering
 \begin{minipage}[b]{0.48\textwidth}
\includegraphics[width=1\linewidth]{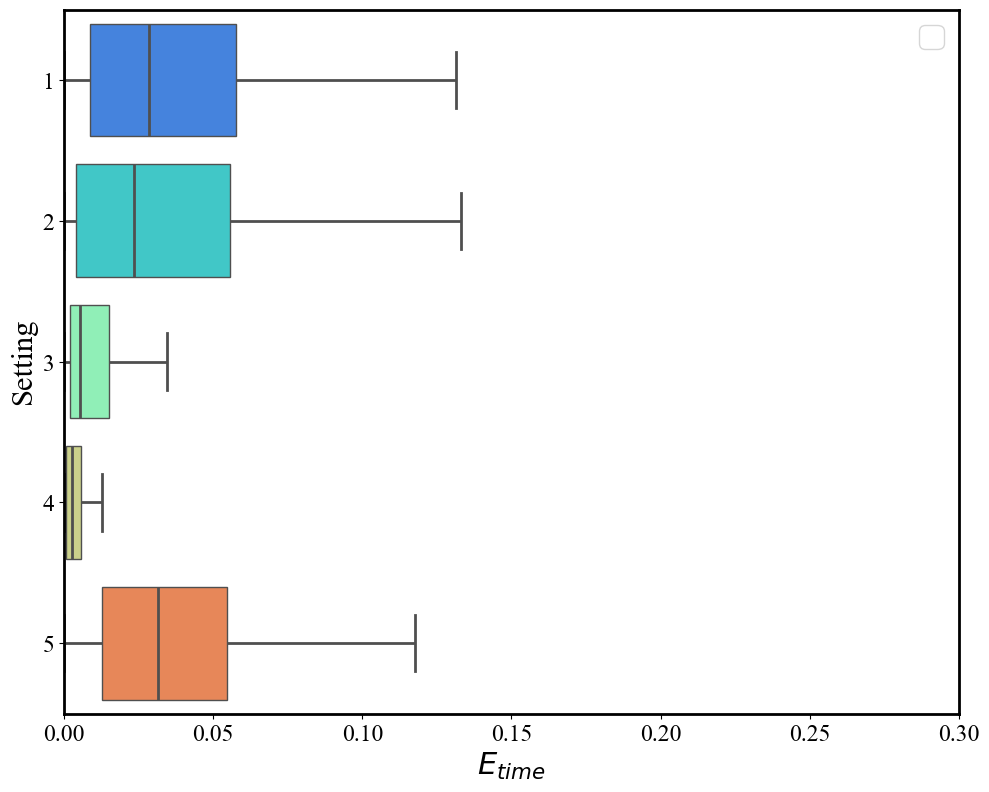}
\end{minipage}}
	\caption{\textbf{Reconstruction result of Pendulum system using polynomial basis with different order:}(a)Absolute error of solution with library parametrized by different order of polynomial basis;(b)Absolute error of time label with library parametrized by different order of polynomial basis.}
	\vspace{0cm}
	\label{fig.poly5}   
\end{figure}

Fig~\ref{fig.poly5} demonstrates that the learned dynamical system, when using the general polynomial basis \(\tmmathbf{\phi}_2\), achieves high precision in reconstructing the solution function (1.2$\%$ RMAE in the noise-free setting) and time labels (0.07$\%$ RMAE in the noise-free setting), albeit with slightly lower performance compared to the larger basis \(\tmmathbf{\phi}_1\) (0.28$\%$ RMAE for the solution function and 0.07$\%$ RMAE for the time labels). However, the Hamiltonian computed along the learned trajectory with the restricted basis exhibits significantly greater variance than that obtained with \(\tmmathbf{\phi}_1\), particularly at higher noise levels. This outcome is expected, as the larger library enables the algorithm to accurately identify the components constituting the conservative system, whereas the polynomial basis does not preserve the Hamiltonian structure, despite yielding an acceptable approximation of the solution function.
To assess the approximation capability of our method relative to the library scale, we approximate the source term using polynomial basis functions of five distinct orders. Fig~\ref{fig.poly5} compares the absolute errors in both the solution function and the time labels for the different bases. As the polynomial order increases from 1 to 4, our method’s results progressively converge toward the ground truth, with optimal performance at order 4—yielding an RMAE of $0.44\%$ for the solution and $0.08\%$ for the time labels. For the Poly(5) basis, however, the sparse regression fails to converge, likely due to the increased sensitivity of higher-order parameters. Moreover, during the parameter identification phase, numerical instability of the forward solver hampers accurate parameter estimation, particularly given the non-uniform sampling of time labels. Consequently, both the solution and time label errors increase.

%% file: conclusion.tex
\section{Conclusion and future work}
\label{sec:conclusions}
Point cloud data lacking temporal labels is a common data type for scientific research. Unveiling the hidden dynamics and reconstructing the missing time label is helpful for understanding the physical laws behind the observation. In this study, we assume that the observation time instants is sampled from a known distribution and the observation data are generated by a ODE system with perturbation.and formulate the reconstruction into a rigorous minimization problems with respect to a continuous transformation of random variable. We perform theoretical analysis of the  uniqueness of smooth transformation given the distribution of observation data and time instants under certain conditions.

In computational perspective, we propose a two-phase learning algorithm to simultaneous estimate the potential trajectory and the hidden dynamical system. In distributional matching phase, we leverage the sliced Wasserstein distance to construct a neural network approximation of the data trajectory and utilize Alternating Direction Optimization technique to distill sparse parameter estimation of the ODE system from a large expression library. In parameter identification phase, we adopt the FSNLS paradigm to further refine the estimation. Based on the empirical observation that complex trajectory renders higher non-convexity and numerical issue for the learning task, we leverage the manifold hypothesis to propose a clustering based partition algorithm to transform long trajectory in to short trajectory segments before training. 

On the experimental side, We demonstrate our method on a number of illustrative and complex dynamical systems exhibiting
challenging characteristic (e.g., chaotic, high-dimensional, nonlinear). Results highlight that the approach is capable of uncovering the hidden dynamics of the real-world point cloud data, and reconstruct the time label in high accuracy.  The proposed method maintains
satisfactory robustness against different types of distribution of observation time instant  (Uniform and truncated Gaussian) and high level 
noises when multiple short trajectories are provided, while in single trajectory setting the perturbation bring about overlapping and lead to failure in trajectory segmentation and time label reconstruction. Results show that the approximation capability is increasing as the library is enlarging, obtaining  finer reconstruction of solution function and time label. However, including high-order terms in the library may introduce numerical instability in the forward computation and the regression process.

As future work, we will try the generalize our method to more challenging regime.  For example, one of the important directions is to unveiling physical laws from PDE-driven or SDE-driven observation without time label. Another interesting direction which is worth exploring is to embed more statistical insight in the algorithm to enhance the robustness for highly perturbed data.

%% file: appendix.tex
\section{Experiment Detail}
In the numerical experiments presented in Sec.~\ref{sec:e}, we generated the reference solution and computed the trajectories of the identified dynamical system using a Runge–Kutta 4 (RK4) scheme on a uniform time grid \(\{t_i^\star\}_{i=0}^{M}\), subsequently evaluating \(\boldsymbol{E}_{\text{sol}}\). Here, we set \(M = 2500\), thereby ensuring that \(dt < \frac{1}{100}\) for all cases and maintaining computational accuracy.

For the observational data of a single long trajectory, we employed agglomerative clustering for trajectory segmentation, with the algorithm implemented using SciPy\cite{2020SciPy-NMeth}. Tables~\ref{tab:supp1} and \ref{tab:supp2} summarize the parameters used in the implementation. Different experiments will have different parameter choices depending on the difficulty of the problem.

\begin{table}[htbp]
    \centering
    \begin{tabular}{ccccccc}
        \toprule
        Parameters & $n_{cluster}$ & library & $\alpha\lambda_0$ & $lr$ & $\lambda_{\text{init}}$ &$\lambda_{\text{Phy}}$ \\
        \midrule
        Linear2D & 10 & Poly(3)+Exp & 0.04 & $3e-4$ & 0.5 & $1e-3$  \\
        Cubic2D & 10 &  Poly(3)+Exp & 0.06 & $6e-4$ & 0.5 & $3e-5$ \\
        Linear3D & 10 &  Poly(3)+Exp & 0.04 & $3e-4$ & 0.5 & $1e-3$ \\
        Lorenz & 10 & Poly(3)+Exp & 0.04 & $3e-4$ & 0.5 & $3e-5$  \\
        LV4D & 10 &  Poly(2)+Exp & 0.045 & $6e-4$ & 0.5 & $3e-4$ \\
        Duffing & 10 &  Poly(3)+Exp & 0.04 & $3e-4$ & 0.5 & $3e-4$ \\
        Pendulum & 10 &  Poly(3)+$\sin$+$\cos$ & 0.04 & $3e-4$ & 0.5 & $3e-4$ \\
        Pendulum$^{\star}$ & 10 &  Poly(3) & 0.02 & $3e-4$ & 0.5 & $3e-4$ \\
        \bottomrule
    \end{tabular}
    \caption{Implement parameters for the numerical examples.$n_{cluster}$ denotes the cluster number in the trajectory segmentation step, Poly(n) represents the n-th order complete polynomials library in d-dimensional space($C_{p+d}^p$ elements), $\alpha\lambda_0$ is the parameter of the proximal operator in the parameter identification phase, $lr$ is the learning rate for the two phase, $\lambda_{\text{init}}\lambda_{\text{Phy}}$ is the weight of Initial Condition loss and Physics loss in distribution matching phase.}\label{tab:supp1}
\end{table}

\begin{table}[htbp]
    \centering
    \begin{tabular}{cccccccc}
        \toprule
        Parameters & B & m & $\lambda_{\text{STRidge}}$ &$\text{Tol}_{\text{STRidge}}$& $Iter_{\text{Phase1}}$ & $Iter_{\text{Phase2}}$ & $Iter_{\text{update}}$\\
        \midrule
        Linear2D & 200 & 100 & 0 & 0.2 &  2000,3000 & 100,500 & 200\\
        Cubic2D & 200 &  100 & 0 & 0.2 & 2500,4500 & 100,650 &200\\
        Linear3D & 200 &  100 & 0.1 & 0.2 & 4500,5200   & 100,850 &200\\
        Lorenz & 200 & 100 & 0.35 & 0.6 & 10000,13200 & 100,1200  &200\\
        LV4D & 200 &  100 & 0.8 & 0.6 & 7500,15000 & 100,1000 &200\\
        Duffing & 200 &  100 & 0 & 0.05 & 2500,4000 & 100,1200 &200\\
        Pendulum & 200 &  100 & 0.6 & 0.6 & 6000,12000 & 100,450 &200\\
        Pendulum$^{\star}$ & 200 &  100 & 0.6 & 0.6 & 6000,12000& 100,450 &200\\
        \bottomrule
    \end{tabular}
    \caption{Implement parameters for the numerical examples.$B$ denotes the batchsize for both two phases, $m$ denotes the direction numbers for the SWD calculation, $\lambda_{\text{STRidge}},\text{Tol}_{\text{STRidge}}$ represent the penalty parameter and the tolerance parameter in STRidge, $(a,b) $ in $Iter_{\text{Phase1}}$ represent the warm up step and the total step in distribution matching phase, $(a,b) $ in $Iter_{\text{Phase2}}$ represents the warm up step and the total step in parameter identification phase. The STRidge updates the parameters every $Iter_{\text{update}}$ step.}\label{tab:supp2}
\end{table}

\section{STRidge Algorithm}
Here we provide the details of the sequential thresholded ridge regression (STRidge) algorithm. In the STRidge method, each linear regression step retains the variables that were not sparsified in the previous regression. And if the original linear equation has $n$ unknowns, the sparse regression operation is performed for a maximum of $n$ iterations. STRidge will terminate directly if either of the following two conditions is met: 1) After a regression step, no additional variables are removed compared to the previous regression; 2) All variables have been removed. For further details of the STRidge algorithm, please refer to Algorithm \ref{alg:ST}.
\begin{algorithm}
\caption{STRidge Algorithm for Solving Linear System $A\mathbf{x}=\mathbf{b}$} \label{alg:STRidge}
\label{alg:ST}
\begin{algorithmic}
\STATE{\textbf{Input}: Coefficient matrix $A \in \mathbb{R}^{m \times n}$, vector $b \in \mathbb{R}^m$, regular terms $\lambda > 0$, threshold $\eta>0$}
\STATE{Compute $x$ by ridge regression $x=\arg\min_{w}\|A\mathbf{w}-\mathbf{b}\|^2+\lambda\|\mathbf{w}\|^2$, set $p=n$;}
\WHILE{True}
\STATE{Select index set $S^+=\{x>\eta\}$, $S^-=\{x\leq\eta\}$;}
\IF{$card\{S^+\}=p$}
\STATE{break}
\ELSE
\STATE $p=card\{S^+\}$ 
\ENDIF
\IF{$card\{S^+\}=0$}
\STATE{break}
\ENDIF
\STATE{$x[S^-]=0, x[S^+]=\arg\min_{w}\|A[:, S^+]\mathbf{w}-\mathbf{b}[:, S^+]\|^2+\lambda\|\mathbf{w}\|^2$}
\ENDWHILE
\IF{$S^+\neq\emptyset$}
\STATE{$x[S^+]=\arg\min_{w}\|A[:, S^+]\mathbf{w}-\mathbf{b}[:, S^+]\|^2$}
\ENDIF
\RETURN 
Vector $\mathbf{x} \in \mathbb{R}^n$ s.t. $A\mathbf{x} \approx \mathbf{b}$
\end{algorithmic}
\end{algorithm}